\documentclass[final]{cvpr}

\usepackage{times}
\usepackage{epsfig}
\usepackage{graphicx}
\usepackage{amsmath}
\usepackage{amssymb}
\usepackage{color}
\usepackage{makecell}

\usepackage[pagebackref=true,breaklinks=true,colorlinks,bookmarks=false]{hyperref}

\def\cvprPaperID{1488} 
\def\confYear{CVPR 2021}

\begin{document}

\title{Image Translation via Fine-grained Knowledge Transfer}

\author{
Xuanhong Chen$^{1}$, Ziang Liu$^{1}$,
Ting Qiu$^{1}$,
Bingbing Ni$^{1,}$\thanks{Corresponding author.}, \\
Naiyuan Liu$^{2}$, Xiwei Hu$^{1}$, Yuhan Li$^{3}$\\
$^{1}$Shanghai Jiao Tong University,
$^{2}$University of Technology Sydney, 
$^{3}$Xian Jiao Tong University\\
$^{1}${\tt\small\{chen19910528,acemenethil,776398420,nibingbing,huxiwei\}@sjtu.edu.cn},\\
$^{2}${\tt\small naiyuan.liu@student.uts.edu.au},
$^{3}${\tt\small colorfulliyu@gmail.com}
}

\maketitle

\begin{abstract}
    Prevailing image-translation frameworks mostly seek to process images via the end-to-end style, which has achieved convincing results.
    Nonetheless, these methods lack interpretability and are not scalable on different image-translation tasks (e.g., style transfer, HDR, etc.).
    In this paper, we propose an interpretable knowledge-based image-translation framework, which realizes the image-translation through knowledge retrieval and transfer.
    In details, the framework constructs a \textbf{plug-and-play} and \textbf{model-agnostic} general purpose knowledge library, remembering task-specific styles, tones, texture patterns, etc.
    Furthermore, we present a fast ANN searching approach, Bandpass Hierarchical K-Means (BHKM), to cope with the difficulty of searching in the enormous knowledge library.
    Extensive experiments well demonstrate the effectiveness and feasibility of our framework in different image-translation tasks.
    In particular, backtracking experiments verify the interpretability of our method.
    Our code soon will be available at \url{https://github.com/AceSix/Knowledge_Transfer}.
\end{abstract}

\section{Introduction}

\begin{figure}[t]
\begin{center}
\includegraphics[width=1\linewidth]{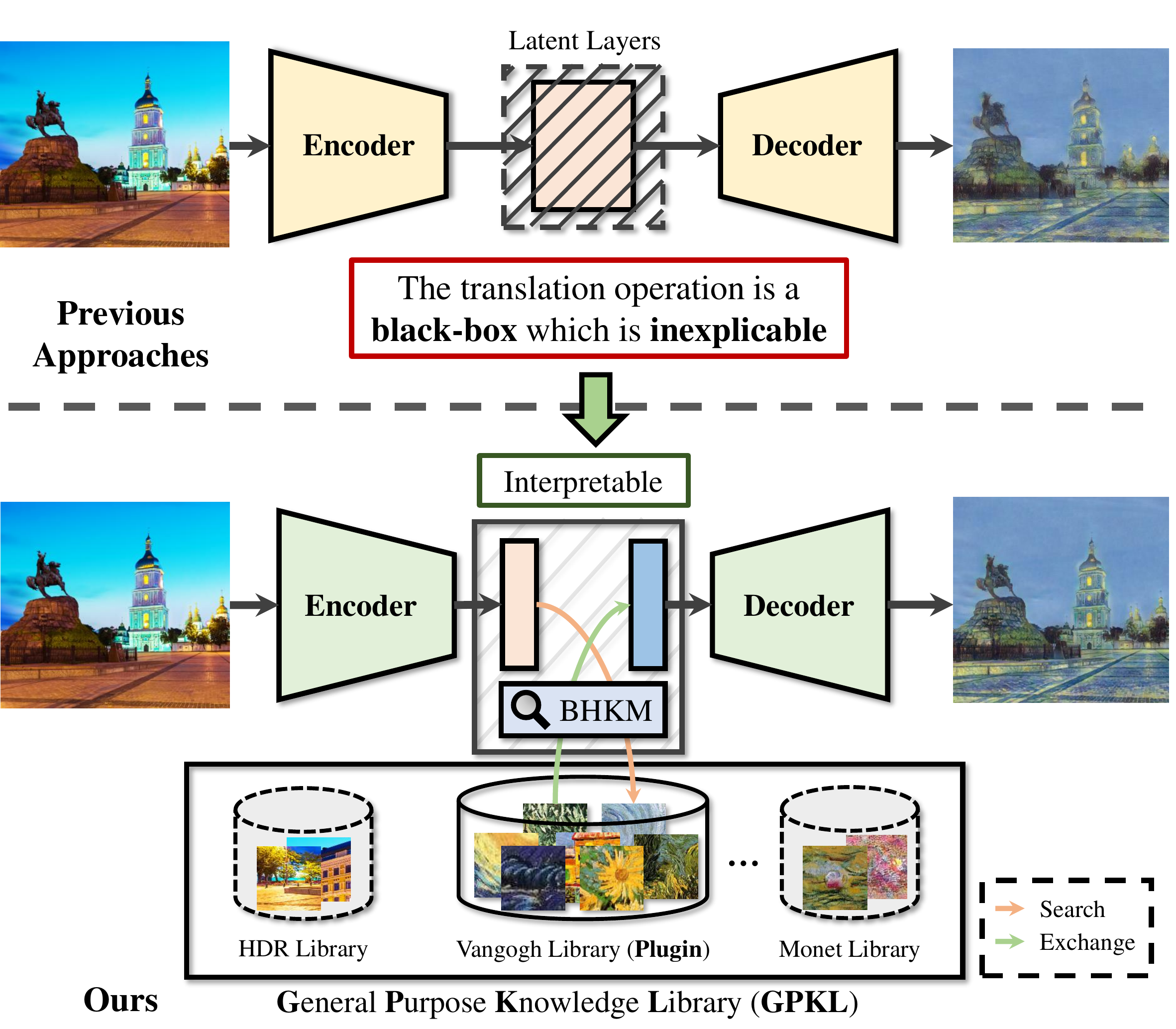}
\end{center}
   \caption{Our proposed image translation approach is an interpretable framework which presents an explicit working logic.
   Cooperated with a plug and play knowledge library, our framework is able to switch in different image translation tasks (\eg, style transfer, HDR, \etc.) without training or fine-tuning.}
\label{fig:motivation}

\end{figure}
\begin{figure*}[ht]	
    \centering	
	\includegraphics[width=1\textwidth]{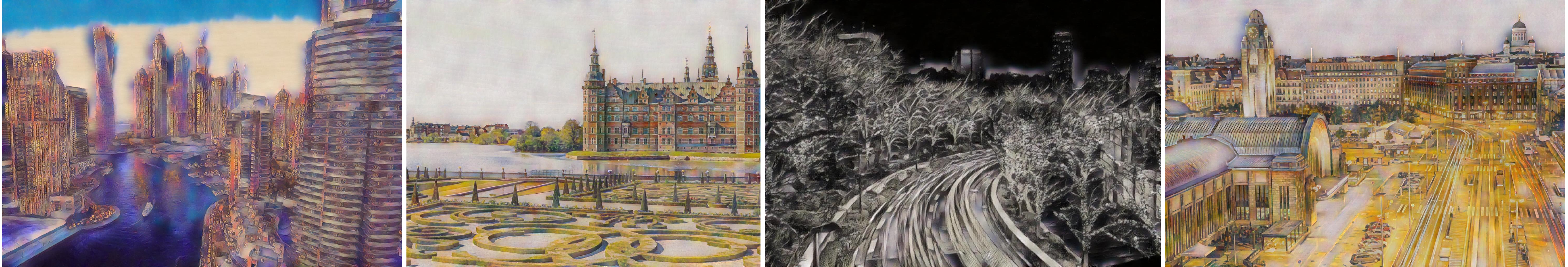}
	\caption{Some translation results generated by our framework. More high-resolution results can be found in suppl.}
    \label{fig:show-off}
\end{figure*}

Image-Translation (\emph{e.g.}, \emph{style transfer} artistically re-renders a target photograph with a given artworks/gallery style, \emph{HDR}, \emph{etc.}) is a promising but challenging computer vision task.
Various CNN-Based approaches~\cite{gatys2015neural,DBLP:journals/corr/ChenS16f,DBLP:conf/cvpr/GuCLY18,sheng2018avatar,chen2017stylebank,DBLP:conf/eccv/JohnsonAF16,DBLP:conf/iccv/SelvarajuCDVPB17,DBLP:conf/mm/ChenYLQN20,DBLP:conf/mm/ChenCNG20} devote their effort into improving the visual effect or reducing the spatio-temporal complexity of translation network.
Although pleasing performance has been achieved, the working mechanism of neural networks are uncertain, which has become the biggest obstacle to deploy CNNs in industrial applications.
Witnessing such a situation, several preliminary works~\cite{DBLP:journals/pami/JegouDS11,fivek,anoosheh2018combogan,DBLP:conf/eccv/ZeilerF14,DBLP:conf/cvpr/MahendranV15,DBLP:conf/iccv/FongV17,DBLP:conf/iccv/SelvarajuCDVPB17,DBLP:conf/iclr/ZintgrafCAW17} 
try to explain the partially internal mechanism, or design a toolbox to visualize the intermediate features for classification task.
However, there are few works invested in designing the interpretable image-translation frameworks.

The most widely used image-translation frameworks are \emph{Auto-Encoder}~\cite{DBLP:conf/cvpr/YaoRX0LW19,DBLP:conf/cvpr/GuCLY18,DBLP:conf/eccv/JohnsonAF16,DBLP:journals/corr/ChenS16f,DBLP:conf/iccv/HuangB17} and \emph{Generative Adversarial Networks (GANs)}~\cite{DBLP:conf/mm/ChenYLQN20,DBLP:journals/corr/abs-2011-01563,DBLP:conf/mm/ChenCNG20,DBLP:conf/eccv/SanakoyeuKLO18,DBLP:conf/iccv/KotovenkoSLO19}, \emph{etc}.
The backbone of these frameworks usually consists of Encoder, Latent-Layers, Decoder structures.
The framework first maps the image into the high-dimensional feature space, and then manipulates these feature maps in hidden layers to realize the translation of input image.
Operators such as conditional injection~\cite{DBLP:conf/cvpr/Park0WZ19,DBLP:conf/iccv/HuangB17,DBLP:conf/iclr/DumoulinSK17}, residual block~\cite{DBLP:conf/cvpr/HeZRS16}, \etc, are usually employed in latent layers to manipulate the feature maps.
Unfortunately, the specific working mechanism of the hidden layers are uncertain, such as why changing the feature map in face editing~\cite{DBLP:journals/corr/abs-2011-01563} can accurately control the closure of the mouth.
For another example, we usually hope that the decoder only focuses on reconstructing the feature maps without participating in the editing, but in fact the decoder greatly affects the effectiveness of editing.
Although this uncertainty does not affect the performance at the current stage, it seriously hinders the framework design and deployment of the model.
For instance, the GAN-Based super-resolution models~\cite{DBLP:conf/cvpr/LedigTHCCAATTWS17,DBLP:conf/eccv/WangYWGLDQL18} have made significant progress in visual effects, but they often produce some undesirable textures, which makes them unreliable.

In this paper, we propose an interpretable image-translation framework that edits images with explicit working logic and procedures.
The most common method used by humans when editing paintings is to find a painting/image that is similar to the current drawing and imitate it.
Empowered by this observation, we design our framework as a knowledge extraction, storage and retrieval system.
Here, we define the terminology \emph{knowledge} used in this paper to represent descriptors (\emph{e.g.}, SIFT, GIST, CNN features, \etc.) of an image.
Unlike the aforementioned frameworks that directly learns an image-to-image mapping,
the establishment of our model includes three main steps:

1). Offline \emph{General Purpose Knowledge Library (GPKL)} establishment,
imitating humans to memorize relevant knowledge in the brain,
we extract knowledge from all images in the target domain dataset and archive them as a database (\emph{i.e.}, GPKL).
The knowledge library is \emph{model-agnostic}, which means that it can be \emph{Plug-and-Play} for the different translation tasks (\emph{e.g.}, style transfer, HDR, \etc.);

2). Offline \emph{Index} Generation,
as GPKLs generally contain a huge amount of data (\emph{e.g.}, more than $1$ million vectors), brute-force search is prohibitively expensive.
To address this challenge, we propose a fast Approximate nearest neighbor (ANN) search algorithm, dubbed \emph{Bandpass Hierarchical K-Means (BHKM)}.
BHKM is a task-specific Hierarchical K-Means~\cite{arai2007hierarchical} algorithm, 
which is based on the following observation:
similar textures are also similar in different frequency bands.
Furthermore, we found that accurate results can be obtained with the matching of partial frequency band.
Our BHKM performs spectral decomposition (\ie, wavelet) on the lowest level elements of hierarchical clustering, and then selectively matches a small part of the bands to reduce the search space.
Once GPKLs are built, we generate the index system for our BHKM;

3). Online image-translation,
for the mentioned frameworks, they directly inference the results via the trained model with knowing NOTHING about what was exactly done on the latent features.
In contrast, our model first extracts the descriptors via a deep model (\emph{e.g.}, VGG, Resnet, \emph{etc.}).
Then, it matches the most suitable knowledge from \emph{Plugged GPKL} according to the geometric characteristics of the descriptors, and apply the knowledge to the descriptors.
Finally, decoder reconstructs the processed feature and produces the final result image.
It can be seen that the whole translation process of our framework has explicit logic and determinate working mechanism.
As our best knowledge, we are the first to try to build an interpretable image-translation framework.
Extensive experiments comprehensively demonstrate that our framework 
is able to produce pleasing translated results in different tasks (\eg, style transfer, HDR, \etc.).
At the same time, the experiments verify that our framework can switch arbitrarily without training on different tasks.
We conduct a backtracking experiment to illustrate the specific translation process of an image to verify the interpretability of our framework.
We can clearly know where the style, tone, \etc of each part come from.
Finally, experiments verify the timeliness and accuracy of the proposed BHKM.

\section{Related Work}
\textbf{Image-Translation.} 
The neural image-translation algorithm was first proposed by Gatys et al.~\cite{gatys2016image,gatys2015texture}, who observed that a pre-trained deep neural network is able to extract multi-level deep features from an arbitrary photograph.
Feed-forward generator networks~\cite{DBLP:conf/eccv/JohnsonAF16,ulyanov2016texture,ulyanov2016instance,chen2017stylebank} were presented to solve the optimization problem. Sanakoyeu et al.~\cite{DBLP:conf/eccv/SanakoyeuKLO18} made further efforts to achieve high-resolution image-translation by proposing a style-aware content loss. Chen and Schmidt~\cite{DBLP:journals/corr/ChenS16f} put forward the first arbitrary image-translation by matching the content patch with the closest style patch. 
Based on the deep feature reshuffling, Gu et al.~\cite{DBLP:conf/cvpr/GuCLY18} presented an efficient approach to improve the results by optimizing the loss in feature domain gradually. 
Furthermore, Chen et al.~\cite{chen2017stylebank} learned the corresponding content and style information effectively by separating network components. 
Zhang et al.~\cite{lu2019closed} was the first to consider the neural image-translation as an optimal transport problem.
Although their results for arbitrary style are appealing, we expect a more efficient and accurate way. 
\begin{figure*}[ht]	
    \centering	
	\includegraphics[width=1\textwidth]{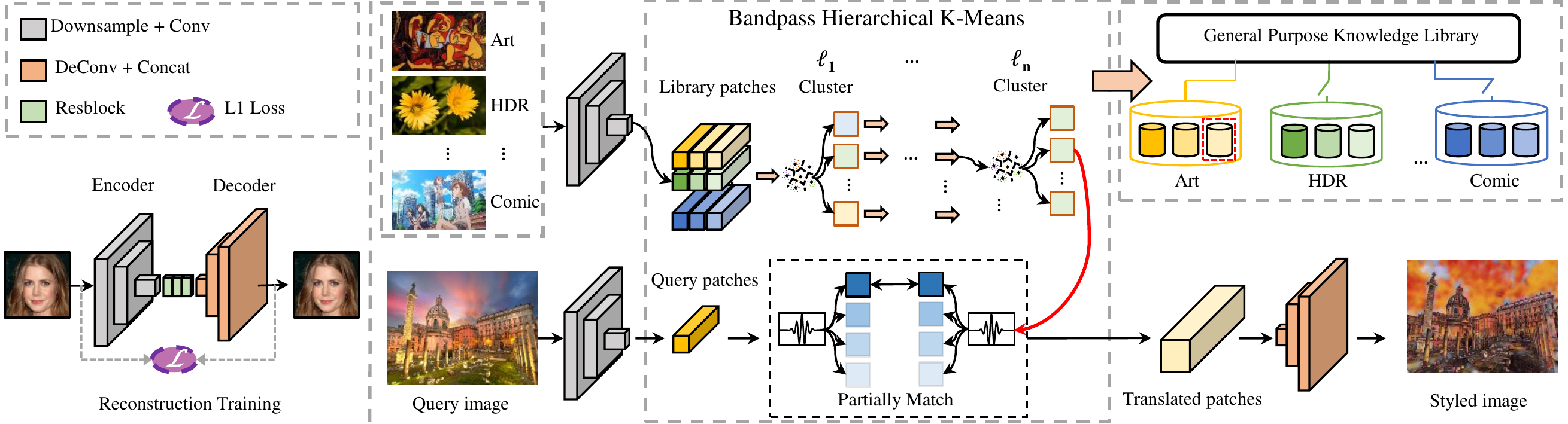}
	\caption{Our framework consists of the Encoder-Decoder, the General Purpose Knowledge Library and the \emph{Bandpass Hierarchical K-Means (BHKM)}.}
    \label{fig:framework}
\end{figure*}

\textbf{Interpretability of CNN.} 
Diverse theoretical work has been proposed to explain how neural network works, including 1). The visualization of CNN representations, and 2). Interpreting semantic knowledge in CNN.
The most direct way of exploring the pattern hidden inside a neural unit is visualization. Zeiler et al.~\cite{DBLP:conf/eccv/ZeilerF14} and Mahendran et al.~\cite{DBLP:conf/cvpr/MahendranV15} estimate the learned feature by inverting them,
while Fong et al.~\cite{DBLP:conf/iccv/FongV17}, Selvaraju et al.~\cite{DBLP:conf/iccv/SelvarajuCDVPB17} and Zintgraf et al.~\cite{DBLP:conf/iclr/ZintgrafCAW17} have considered visualizing the regions which are important for decision. 
Interpreting CNN semantic knowledge is another intriguing direction. Network dissection~\cite{DBLP:conf/cvpr/ZhangWZ18a,DBLP:conf/cvpr/BauZKO017} measures the overlap between highly activated regions and labeled concepts.
However, almost all the approaches above aim to explain the models for discriminative tasks like image classification, leaving image-translation to be explored.

\section{Methodology}
\subsection{Overview}
People are very confused about the internal working mechanism of the image-translation framework, which makes the framework difficult to deploy and cannot be effectively diagnosed.
To alleviate this problem, we propose an interpretable image-translation framework. 
This framework is proposed based on the following observations.
Imitation is a very important way for humans to create paintings.
By analogy, image-translation can also be achieved by imitating samples with similar geometric shapes in a given data set.
Furthermore, if this data set is large enough, the framework can achieve most of the required translation effects via simple imitation.
As show in the Fig.~\ref{fig:framework}, our framework consists of the Encoder-Decoder, the General Purpose Knowledge Library and the \emph{Bandpass Hierarchical K-Means (BHKM)}.
The Encoder-Decoder is designed to extract the knowledge of input image and reconstruct the translated image.
GPKL is similar to human memory, storing all the discriminative knowledge extracted from the data set.
It is worth mentioning that GPKL is \emph{model-agnostic}, which can be \emph{plug and play} according to the purpose of the task.
Due to the huge volume of GPKL, we propose a task-specific Hierarchical K-Means algorithm BHKM to achieve low temporal-spatial complexity search in GPKL.

\subsection{Image-Translation Problem Formulation}
Image-to-image-translation, an indispensable task in the computer vision field, owns various applications, like image style transfer, attribute manipulation, domain adaptation, etc.
Image-to-image translation learns a mapping from a source domain to a target domain, which converts an image to not be distinguished by the target domain and preserves the main input characters.
Let $\mathcal{F}_{s \mapsto t}$ presents the source-to-target translation function, the image-translation between source-domain images $\mathbb{I}_s= \{\textbf{s}_{1}, \textbf{s}_{2},...\}$ and target-domain images $\mathbb{I}_t= \{\textbf{t}_{1}, \textbf{t}_{2},...\}$ is as follows:
\begin{equation}
\textbf{o}_{ij} = \mathcal{F}_{s \mapsto t}(\textbf{s}_{i},\textbf{t}_{j}),
\label{eq:ImageTranslation}
\end{equation}
where $\textbf{s}_{i}$ is an input image, $\textbf{t}_{j}$ denotes the target image and $\textbf{o}_{ij}$ is the translated result.

\subsection{The General Purpose Knowledge Library}

In order to effectively remember the discriminative image knowledge of the target domain data set, we designed the \emph{General Purpose Knowledge Library (GPKL)}.
A very simple idea is to directly use the pixel information of the image as the knowledge.
But the unprocessed pixel information is noisy and unstable to lighting condition and color tone.
The primary features (\emph{e.g.}, Image Structure Tensor~\cite{bigun1987optimal}, SIFT, \emph{etc.}) of the image can also be selected as candidates, but these feature extraction methods are too simple to effectively extract the semantic information in the image, so that the extracted knowledge has a lot of redundancy and is sensitive to color tone.
In our framework, we employ the CNN (\eg, VGG, Resnet, \etc.) as the knowledge extractor. 
Compared with the primary feature, CNN feature provides rich semantics, it can still produce robust features in different scenarios, which coincides with the general purpose of our architecture.

We first leverage the pre-trained \emph{Encoder} $\mathcal{E}$ to obtain the hidden features $\mathbb{H}_t= \{\textbf{h}_{1}, \textbf{h}_{2},...\}$ from target-domain images $\mathbb{I}_t= \{\textbf{t}_{1}, \textbf{t}_{2},...\}$.
Since the coarse spatial scale of these features, they cannot be used to represent the image knowledge in target-domain dataset.
We further extract patches $\mathbb{A}_t = \{\textbf{a}_{1}, \textbf{a}_{2},...\}$ by sliding a window over features $\mathbb{H}_t$ with window size for $W_h\times W_w$ and stride for $W_s$.
$\mathbb{A}_t$ contains all texture patterns in the target domain,
each cell $\textbf{a}_{i}$ carries the domain knowledge of the corresponding geometric pattern.
Matching the geometrically similar $\textbf{a}_{i}$, then transfer its knowledge into the input image patches, we can translate the input image.
Experiments have found that there is a lot of redundant information in $\mathbb{H}_t$.
For example, $\mathbb{A}_t$ corresponds to Van Gogh's \emph{Road with Cypresses 1890} gallery.
There are many similar brushstrokes in the gallery, these are redundant knowledge.
Given this, we perform the de-redundancy operation on the $\mathbb{A}_t$ to merge redundant knowledge, which also reduces the search space.
We select Normalized Cross-correlation (NCC) for similarity measurement between $\textbf{a}_{i}$. 
According to~\cite{DBLP:journals/corr/ChenS16f}, NCC is an efficient approach to measure high-dimension similarity for nearest neighbor searching.
The NCC similarity between $\textbf{a}_i$ and $\textbf{a}_j$, $i,j=\{1,2,...\}$, is as follows:

\begin{equation}
\mathcal{S}(\textbf{a}_i, \textbf{a}_j)= \frac{\langle \textbf{a}_i, \textbf{a}_j \rangle}{||\textbf{a}_i|| \cdot ||\textbf{a}_j||}.
\label{eq:cossimilarity}
\end{equation}

Once de-redundancy is completed, we have completed the construction of a Knowledge Library $\textbf{L}_i$ .
Collecting knowledge libraries corresponding to different purposes (\eg, style transfer, HDR, \etc.) will finally build the \emph{General Purpose Knowledge Library} $\mathbb{L}= \{\textbf{L}_{1}, \textbf{L}_{2},...\}$.
Deserve to be mentioned, there is a type of patch-based style transfer frameworks~\cite{DBLP:journals/corr/ChenS16f,DBLP:conf/cvpr/GuCLY18}, which realize the style transfer between two images by exchanging similar patches in them.
These approaches can be considered as special forms where the length of our knowledge library $\textbf{L}_i$ is equal to 1.
In fact, the representation ability of the library formed by a single image is incomplete, because the geometric patterns contained in a single image are particularly scarce.


\begin{figure}[!tb]
\centering
\includegraphics[width=0.49\textwidth]{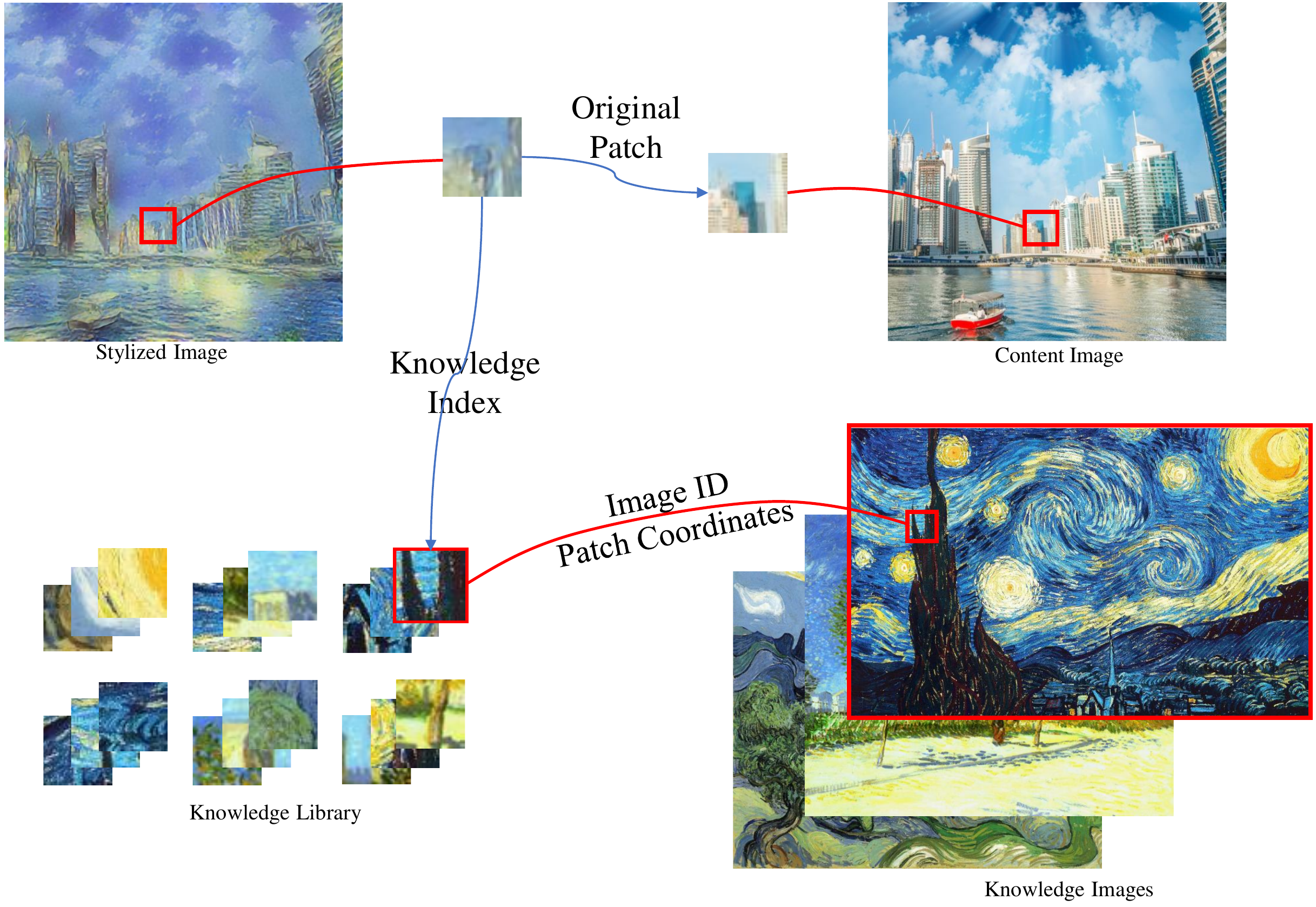}
\renewcommand{\tabcolsep}{42pt}
\caption{Knowledge backtracking pipeline. After a content patch is matched, we can acquire the library index of the corresponding knowledge. This index then leads to the source image and receptive field coordinates of that knowledge.}
\label{fig:retrace}
\end{figure}

\subsection{Matching in the General Purpose Knowledge Library}
In order for GPKL to provide enough knowledge to effectively represent the image attributes (\eg, style, color-tone, \etc.) of the target domain, GPKL is generally very large.
Therefore, we need a fast big data search algorithm that fits our task.
In this section, we will introduce our proposed \emph{Bandpass Hierarchical K-Means (BHKM)} search algorithm to achieve fast matching in our GPKL.
Hierarchical K-Means is a high performance Approximate Nearest Neighbor (ANN) search framework for non-exhaustive search, inverted indexing, \etc, whose logic behind is to continuously subdivide the data into smaller data clusters to continuously reduce the search space.

{\bf The Bandpass Hierarchical K-Means Algorithm.}
Formally, suppose $\textbf{L}$ is a Knowledge Library, suppose that $\textbf{L}$ contains a total of $\textbf{N}$ knowledge vectors, that is $\textbf{L}= \{\textbf{b}_{1},..., \textbf{b}_{i},..., \textbf{b}_{N}\}$ and $\textbf{b}_{i} \in \mathbb{R}^{1\times D}$.
We first use K-means to divide the entire $\textbf{L}$ into $K^{*}_{1}$ sub-clusters to reduce the search space in $1st$ level, and get the sub-sets $\textbf{L}= \{\textbf{L}^1, \textbf{L}^2,..., \textbf{L}^{K^{*}_{1}}\}$, $\textbf{L}^{k^*_1}=\{\textbf{b}^{k^*_1}_{1}, \textbf{b}^{k^*_1}_{2},...\}$.
Let $\mathcal{C}$ represents the center allocations of K-means. $\textbf{B}$ denotes cluster center of a subset.
The clustering optimization potential function is expressed as follows:
\begin{equation}
\min_{\textbf{B}} \min_{\mathcal{C}} \sum_{j=1}^{K^*_1} \sum_{i:\mathcal{C}_i=j} \mathcal{S}(\textbf{B}_j, \textbf{b}_i),
\label{eq:kmeans}
\end{equation}
where 
\begin{equation}
\mathcal{C}^{(t)}_i(\textbf{B}, \textbf{b}_i) = \mathop{\arg\max}_{j=1,2,\dots,K^{*}} \mathcal{S}(\textbf{B}_j, \textbf{b}_i),
\label{eq:center}
\end{equation}
\begin{equation}
\textbf{B}^{(t+1)} = \mathop{\arg\max}_{\textbf{B}} \sum_{i:C_i=j} \mathcal{S}(\textbf{B}^{(t)}_j, \textbf{b}_i),
\label{eq:mu}
\end{equation}
where $t$ represents the number of clustering iterations.
We iteratively perform a K-means clustering on the sub-clusters obtained at the previous level, and finally we will divide $\textbf{L}$ into $K^*_l\cdot K^*_{l-1}$ sub-sets.
In this way, the original high-dimensional problem is reduced to $M$ simple sub-problems.
This is how Hierarchical K-Means works.
Although it has been divided many times, the number of samples in these sub-clusters is still relatively large in the actual scene.
A direct idea is to quantify the sub-cluster using the method of \emph{Product Quantization}~\cite{DBLP:journals/pami/JegouDS11}.
In fact, we have tried this scheme, but for NCC distance, PQ cannot guarantee the optimal solution.
Furthermore, our experiments reveal that the PQ scheme often produces serious artifacts.
Similar textures are also similar in different frequency bands.
Empowered by this observation, we propose to perform spectral decomposition on the last level of sub-sets, and we only match partial components to reduce the computational burden of search.
For example, for some geometric textures, we even only need to pay attention to the low frequency part to get very accurate matching results.
The process of our proposed BHKM is shown in Fig.~\ref{fig:framework}.
We employ the \emph{Wavelet Package Transform (WPT)} to decompose vectors in the sub-sets.
WPT decomposes one patch into a series of smaller wavelet components of the same size, called a wavelet packet.
We employ simplest Haar wavelet as WPT, because it's enough to obtain patches' information and it reduces time consumption.  


\subsection{Image-Translation via Knowledge Transfer}
Humans deal with new things often by recalling familiar things or memory experiences.
Our framework for image processing is to imitate this process.
The just-established knowledge base is like a huge memory warehouse,
and available knowledge can be found in this warehouse for the various geometric patterns contained in the input image.

For a input image $\textbf{I}_s$, we first input it into the Encoder $\mathcal{E}$ to obtain its features, which provides the non-trivial geometric patterns (\eg, eyes, face, \etc.).
Then, by sweeping the same size (\ie, $W_h\times W_w$) window as above, the features are sliced into patches $\mathbb{A}_s = \{\textbf{a}_{1}, \textbf{a}_{2},...\}$.
The stride for sliding can affect the search space and also affect the quality of the translation results.
Small stride size will get smooth generation results, but huge search space.
$\mathbb{A}_s$ contains all geometric patterns that need to be processed.
As mentioned above, we look for the most similar knowledge in the GPKL library for each element in $\mathbb{A}_s$.
For ease of illustration, we take one of patch $\textbf{a}_s$ as an example.
We apply the BHKM to get the candidate set of the nearest neighbor vectors for $\textbf{a}_s$ in the activated Knowledge Library.
Then, we get the nearest neighbor vector for $\textbf{a}_s$ via Top-K.
We transfer the knowledge from $\textbf{a}_t$ to $\textbf{a}_s$ through the knowledge transfer operator, which realizes the editing of pictures.
Then reconstruct the processed feature through the decoder to obtain the translated image.
We provide two means for transferring the information from most similar knowledge.

\textbf{Geometric Swap.} It is first proposed in~\cite{DBLP:journals/corr/ChenS16f}.
Geometric Swap directly replaces source image own patch with the target's patch.
Its specific process can be expressed as the following formula:
\begin{equation}
\textbf{a}_s = \textbf{a}_t.
\end{equation}

\textbf{Statistics Swap.} Inspired by AdaIn~\cite{DBLP:conf/iccv/HuangB17}, we propose a statistics swap, which directly exchanges the mean and variance of two similar patches.
Unlike Adain, our statistics exchange method does not directly exchange statistics for the entire image.
Our operator only exchanges local statistics of similar places, which allows it to maintain the content of the processed image well.
\begin{equation}
\textbf{a}_s = \frac{\textbf{a}_s - \mu(\textbf{a}_s)}{\sigma(\textbf{a}_s)}\cdot\sigma(\textbf{a}_t)+\mu(\textbf{a}_t).
\end{equation}

\subsection{Reconstruction of the Latent Features}
The translation process of our framework is conducted in the feature space.
Actually, the extracted features in CNN network contain a lot of semantic information but little style information~\cite{gatys2015neural}, which is crucial for image translation.
In order to build a friendly feature space for the translation processing, we design a Auto-Encoder (AE) structure, to extract features from content image and reconstruct the processed image from latent space.
We follow the SOTA approaches~\cite{gatys2015neural,DBLP:journals/corr/ChenS16f,DBLP:conf/cvpr/GuCLY18,DBLP:conf/iccv/HuangB17} to build the encoder as a VGG~\cite{luan2017deep} structure, which has proven to be feasible and very effective in image-translation.
For the decoder, its task is to reconstruct the translated image from latent space.
The decoder consists of multiple groups of resize + convolution blocks, which can alleviate the checkerboard artifacts problem caused by transpose convolution.

\textbf{Loss Function.}
Equipped with the GPKLs, the incremental learning in our framework demands no training anymore.
The only part needs to be trained in our framework is the AE.
The reconstruction loss $\mathcal{L}$ of our AE is defined by L1 loss between the input image $I_i$ and the reconstructed image $I_o$:
\begin{equation}
\begin{split}
\mathcal{L}(\mathit{I_o},\mathit{I_i})=
\parallel \mathit{I_o} - \mathit{I_i} \parallel_1.
\end{split}
\label{eq:loss1}
\end{equation}

\begin{figure}
\centering
\includegraphics[width=0.49\textwidth]{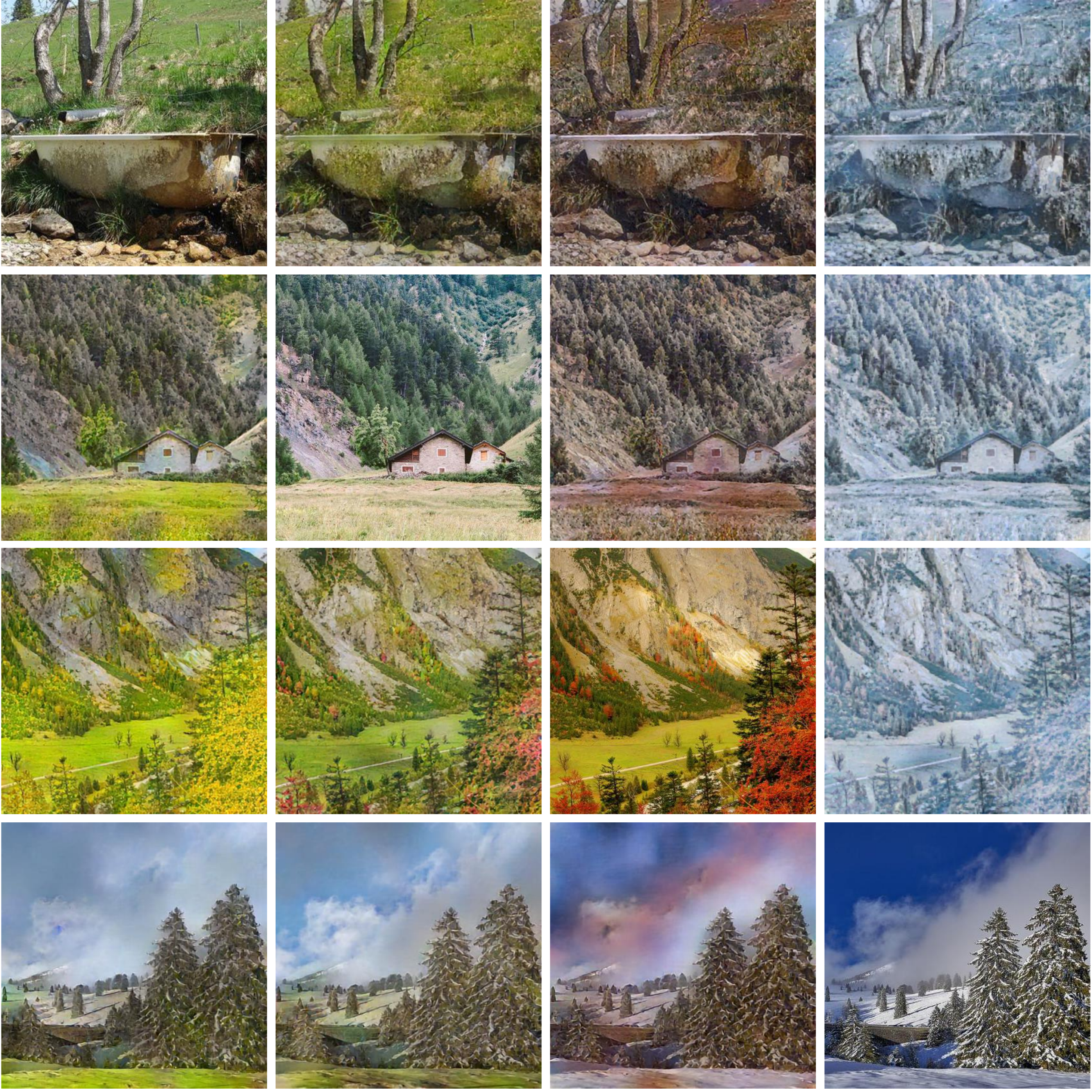}
\setlength{\tabcolsep}{0.031\textwidth}
\begin{tabular}{cccc}
Spring & Summer   &Autumn  &  Winter\
\end{tabular}
\caption{Season transfer results. The images in diagonal positions are original ones and others are generated through knowledge transfer.}
\label{fig:season}
\end{figure}

\section{Experiment}
The purpose of the proposed method is to conduct interpretable image-translation while achieving competitive performance. To verify its effect, we conducted various experiments on several image-translation tasks, including style transfer, HDR, \etc. 
Furthermore, we conduct comparison experiments with previous methods to evaluate the efficiency and capability of our method. 



\subsection{Experiment Setting}

Following the study in~\cite{DBLP:journals/corr/ChenS16f}, our framework chooses 2$\times$2 as the patch size considering a trade-off between the performance and the resource limitation.
Our knowledge library does not require training, while the image reconstruction module demands a trained auto-encoder.
The encoder uses layer $relu3\_1$ of VGG-19~\cite{luan2017deep} network pretrained on ImageNet~\cite{geirhos2018imagenet} dataset.
The decoder contains 5 residual blocks~\cite{DBLP:conf/eccv/JohnsonAF16} and 3 upsampling blocks (\ie, interpolation+convolution). 
All layers above contain a SeLU~\cite{DBLP:conf/nips/KlambauerUMH17} non-linear activation layer. The last upsampling block contains no activation layer and uses bias instead~\cite{DBLP:conf/iclr/KarrasALL18}.

In this work, only the decoder needs training since the well pretrained VGG encoder is used. The decoder is trained with $L1$ reconstruction loss on a diversified dataset combining Place365~\cite{wu2018pair} and Wikiart~\cite{DBLP:conf/iccv/HuangB17,DBLP:conf/iclr/DumoulinSK17}. An Adam optimizer of learning rate 1$e$-4 is applied. 
All experiments are conducted using GTX 1080Ti GPUs with python 3.7 in an Ubuntu system.

\begin{figure}
\centering
\includegraphics[width=0.48\textwidth]{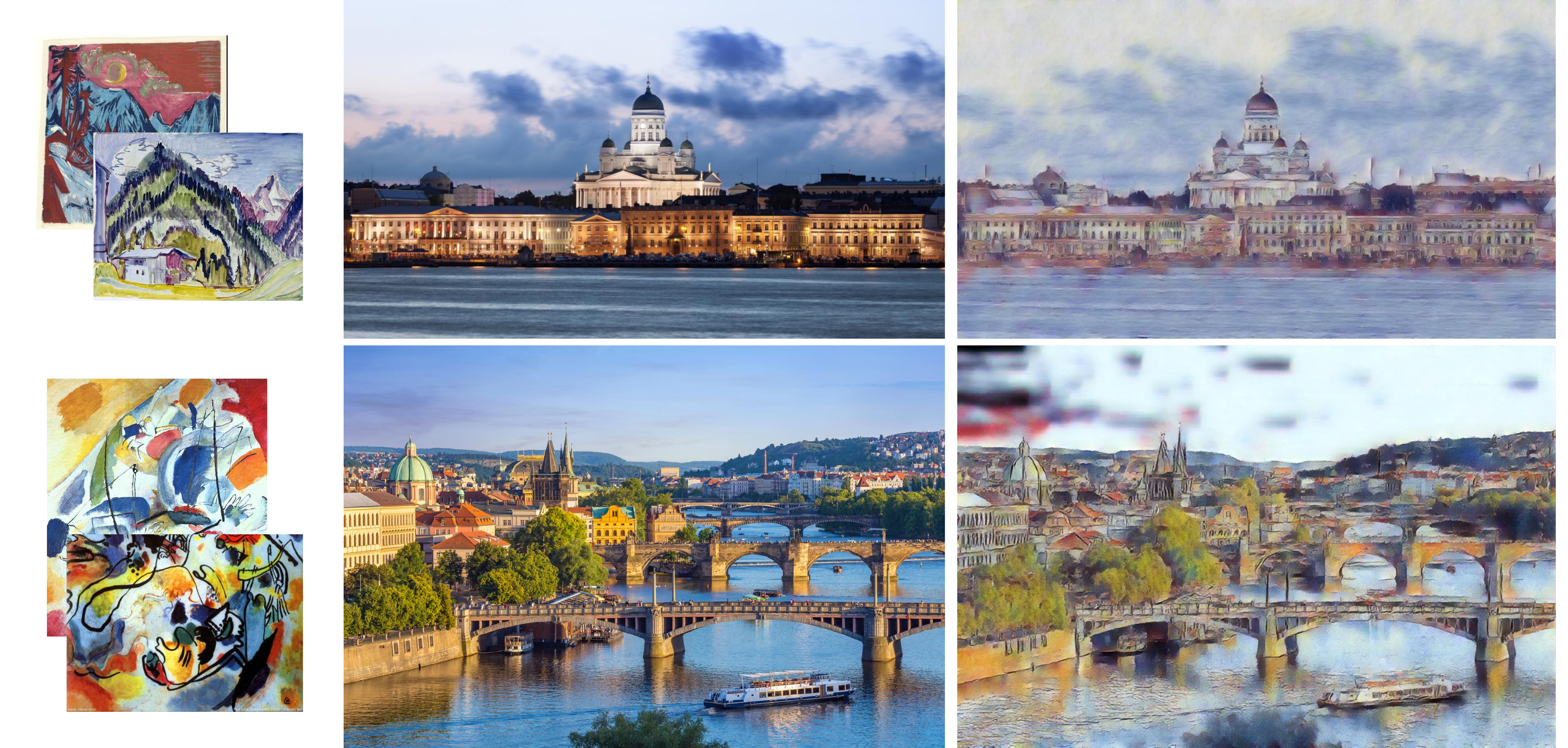}
\caption{Stylization results with libraries containing multiple images.}
\label{fig:style_result}

\centering
\includegraphics[width=0.47\textwidth]{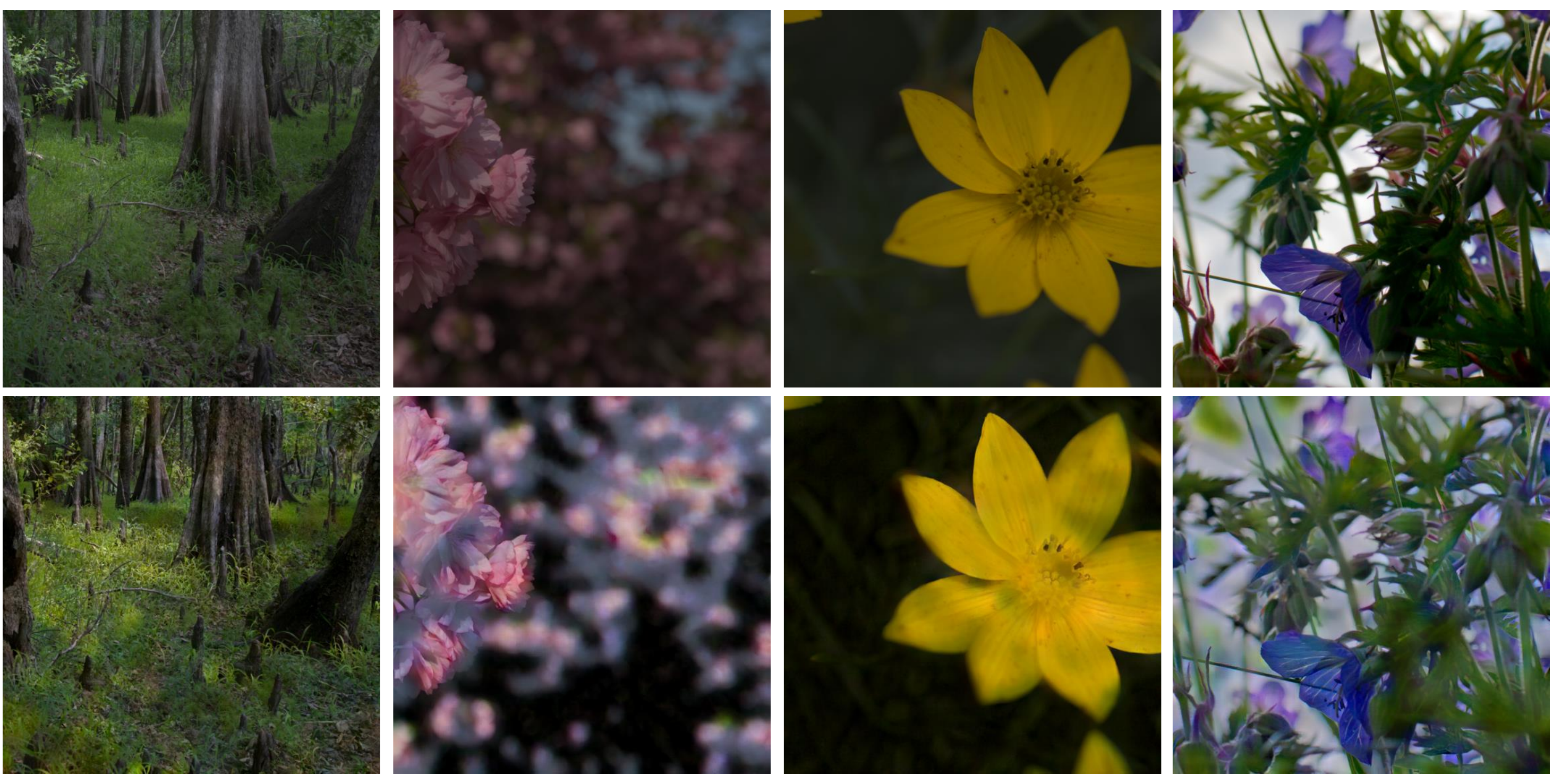}
\caption{HDR images recovered through knowledge transfer. In the first row are original LDR images and in the second row are HDR images recovered from positionally corresponding images.}
\label{fig:HDR_result}
\end{figure}

\subsection{General Performance Analysis}
In this section, we present the generalized image-translation ability of our method by analyzing its performance on various tasks.

{\bf Style Transfer.}
Using geometric swap, our method can fuse the characteristics of numerous style sources on one content image. This ensures the abundance of semantics in stylized results and robust adaptation to diverse contents. In Fig.~\ref{fig:style_result} are two stylized results by our method. Despite the complex structures, our method manages to stylize the contents smoothly and effectively.



{\bf Image HDR.}
Unlike stylization, HDR requires the integrity of semantic details. We propose altering the dynamic range of image through statistic knowledge transfer. Since only regional statistics are shifted, almost no damage to details is involved. We use Mit5K~\cite{fivek} to evaluate our method's performance. The results are given in Fig.~\ref{fig:HDR_result}.

{\bf Season Transfer.}
Another important task of image-translation is Season Transfer. It requires the model to maintain image details while modifying seasonal traits, which can be undertaken by shifting regional feature statistics. We test our method on the season dataset introduced by ~\cite{anoosheh2018combogan}. The results are given in Fig.~\ref{fig:season}.

\begin{figure}
\centering
\includegraphics[width=0.49\textwidth]{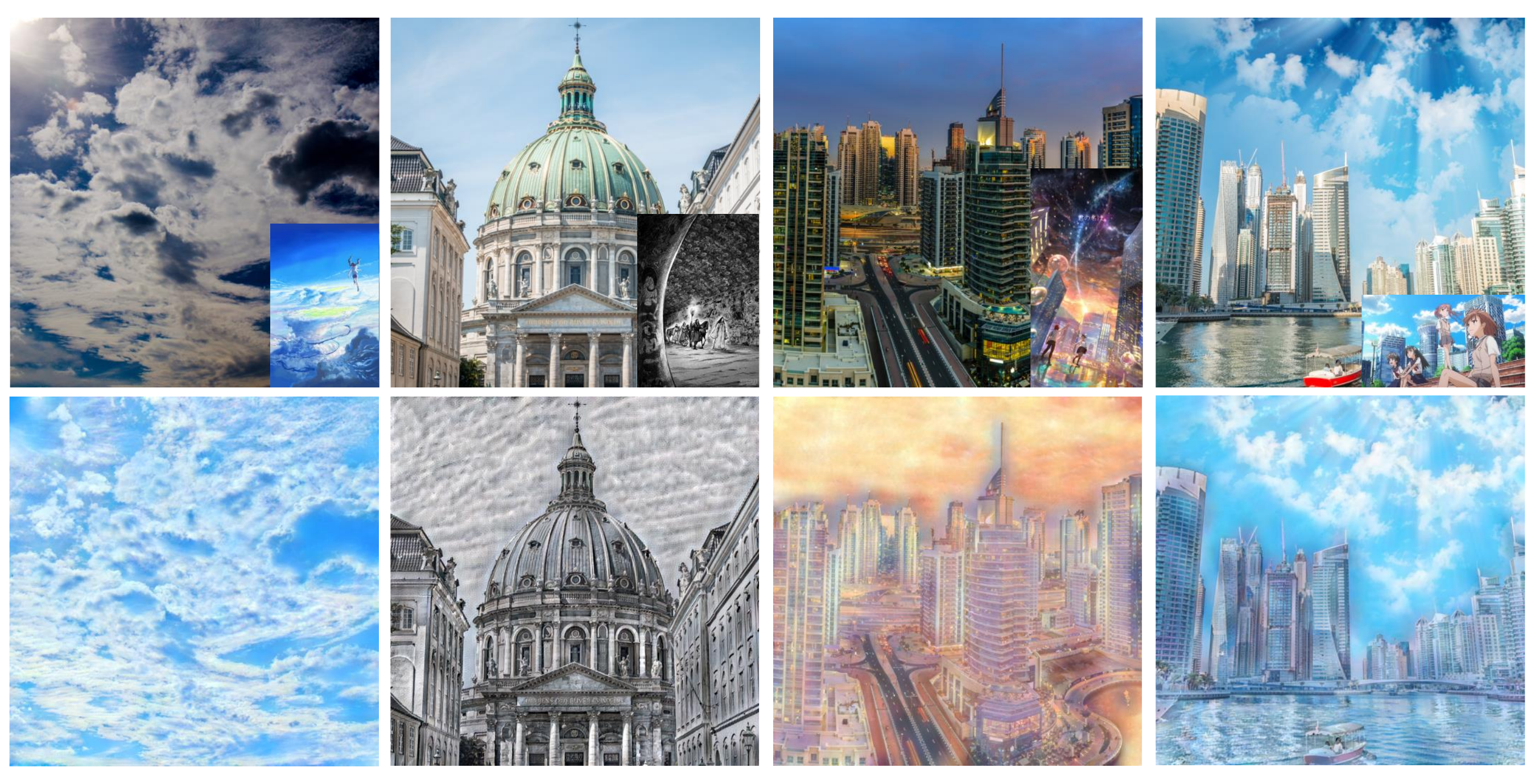}
\caption{Photo to comic results by knowledge transfer. The original content images are placed in the first row with its target comic styles, while the results are show in the second row.}
\label{fig:comic_result}
\end{figure}

{\bf Comic Style Transfer.}
Comic Style Transfer~\cite{DBLP:conf/cvpr/ChenLL18} is a special task in image-translation. It set itself apart from traditional style transfer since comic is a different form of art with unique aesthetic standards. To validate the general image-translation effectiveness of our method, we also conduct comic style transfer with comic images gathered from the internet. Several representative comic or cartoon scenes are selected. The results are given in Fig.~\ref{fig:comic_result}.

\begin{figure*}
\setlength{\tabcolsep}{0.02\textwidth}
\centering
\includegraphics[width=0.99\textwidth]{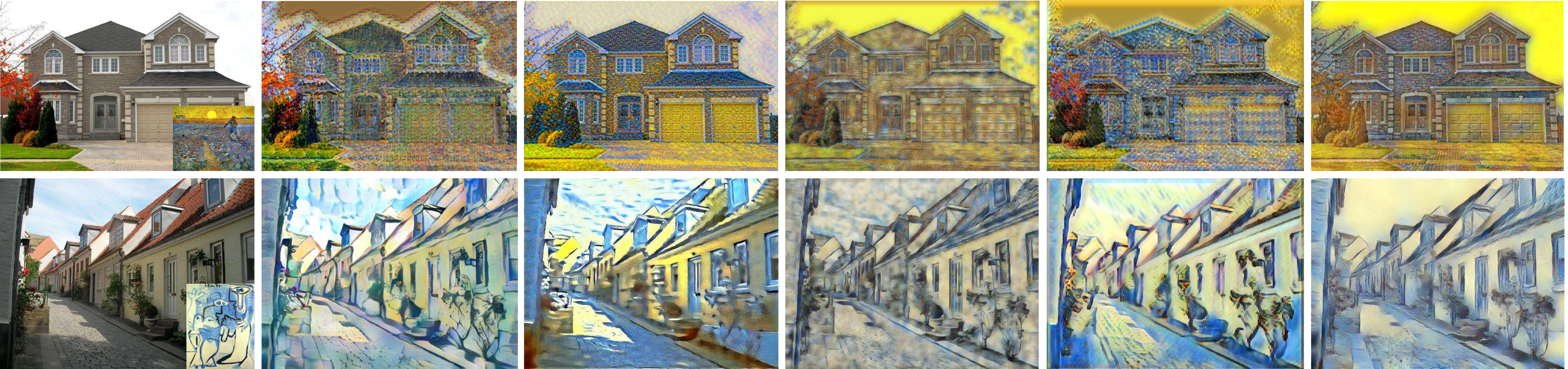}
\setlength{\tabcolsep}{0.013\textwidth}
\begin{tabular}{p{3cm}<{\centering}p{1.7cm}<{\centering}p{3cm}<{\centering}p{2.5cm}<{\centering}p{2.5cm}<{\centering}p{2cm}<{\centering}}
Contents \& Styles & AdaIN~\cite{DBLP:conf/iccv/HuangB17}  & Gatys et.al~\cite{gatys2015neural} &   Patch Swap~\cite{DBLP:journals/corr/ChenS16f} & StyleBank~\cite{chen2017stylebank} &  ours \\
Inference time & 0.011s & 58s & 1.58s  &   0.13s &  0.29s  \\
\end{tabular}




\caption{Visual effect comparison between Patch Swap~\cite{DBLP:journals/corr/ChenS16f}, StyleBank~\cite{chen2017stylebank}, AdaIN~\cite{DBLP:conf/iccv/HuangB17}, method by Gatys et.al~\cite{gatys2015neural} and ours. Our method achieves competitive visual effect while being interpretable.}

\label{fig:Cmp_sota}
\end{figure*}


\subsection{Comparison with previous methods.}
In this section, we compare our method with 4 other methods: 
method proposed by Gatys et al.~\cite{gatys2015neural}, 
Patch Swap~\cite{DBLP:journals/corr/ChenS16f}, AdaIN~\cite{DBLP:conf/iccv/HuangB17}, StyleBank~\cite{chen2017stylebank}
. To compare our method with previous works, we choose style transfer as the comparison task.
We reproduce the performance with their official codes and make an analysis as shown in Fig.~\ref{fig:Cmp_sota}. 

The results show that our method has unique quality among all other models.
Stylebank has similar functions as our method but is burdened with long incremental training time and overfitting problems for every new style. In contrast, it only takes our method several minutes to take on a new style or even task. 
Patch Swap proposed by Chen et al.~\cite{DBLP:journals/corr/ChenS16f} has also strong adaptability and high efficiency, but its results are often over-smooth and makes image textures and colors indistinct.
The method of Gatys et al.~\cite{gatys2015neural} struggles at preserve the local textures as distinct edges, in which our method performs quite better.
Our method is slightly behind AdaIN in performance but avoids halo artifacts and distortion, which AdaIN often encounters.
Experiments prove that our method still has an effect that is not inferior to other methods when our method is interpretable.

\section{Ablation Study}
In this section, we explicitly analyze the interpretability of the proposed method. Then, we exploit the effects of statistics swap and geometric swap. Finally, we compare our BHKM with other methods with similar functions.

\subsection{Interpretability Analysis}
An essential attribute of our method is interpretability. Every swapped knowledge can be backtracked. By finding the most similar knowledge in the library we can also acquire its corresponding image and receptive field coordinates. This information is stored along with source patches in the knowledge library construction stage. The backtracking pipeline is graphically explained in Fig.~\ref{fig:retrace}.


Such an attribute not only enables the explicit understanding of image-translation process, but also provides the means for users to deliberately manipulate image generation. To be exact, the construction of knowledge library can be concretely designed for specific tasks and even individual images. Knowing the semantics and structures of content images, it is convenient to construct a library with images that are more likely to contain the matching features. Even a small alteration to library composition can benefit the result significantly. We conduct experiments of HDR task to reinforce this presumption.

\begin{figure}[!tb]
\centering
\includegraphics[width=0.49\textwidth]{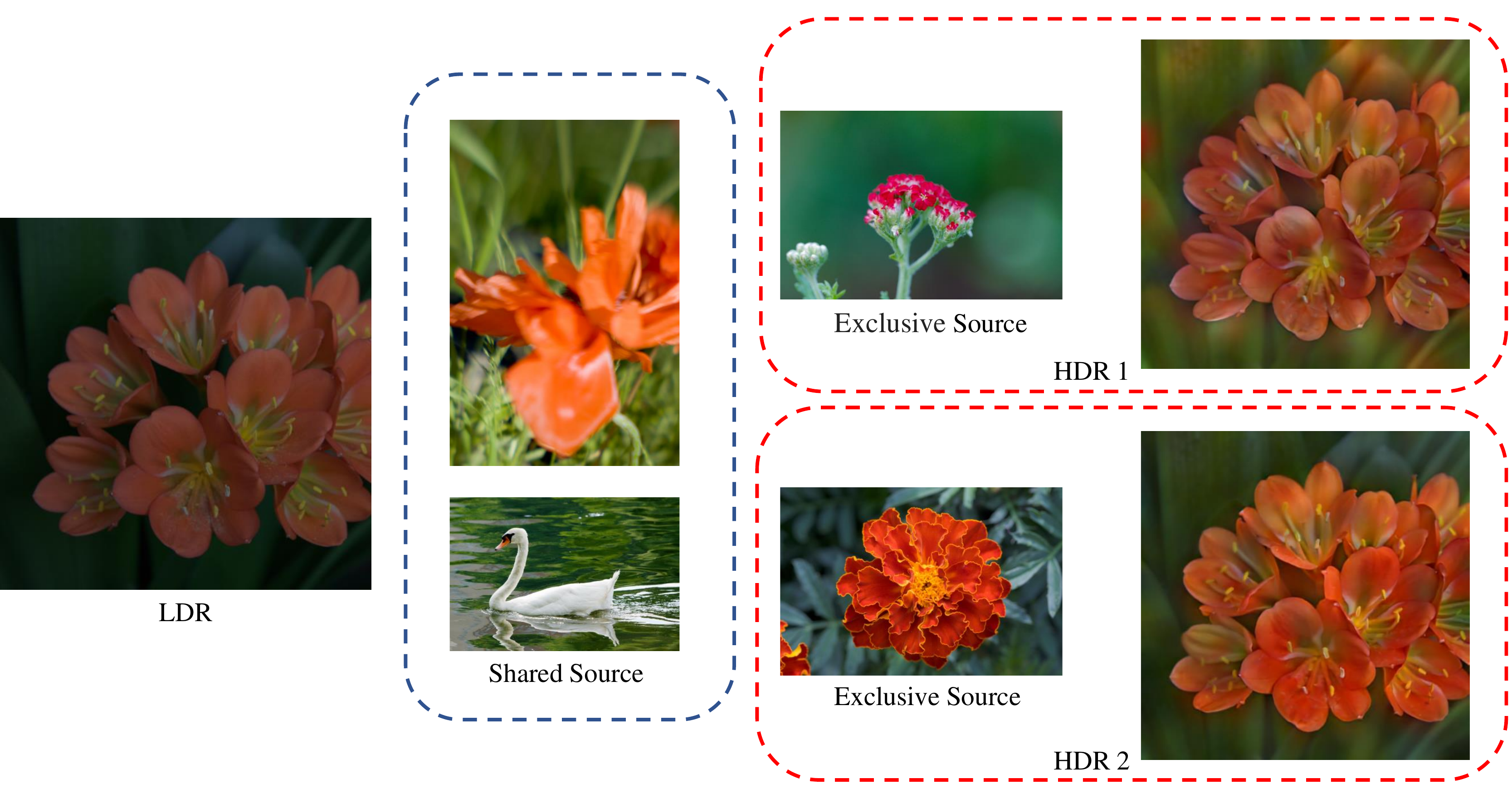}
\renewcommand{\tabcolsep}{42pt}
\caption{HDR images using two different libraries. Inside the blue frame are source images shared by both libraries. Inside the red frames are the exclusive source image and the HDR result for either library.}
\label{fig:hdr_compare}
\end{figure}

As is seen in Fig.~\ref{fig:hdr_compare}, the HDR result from the first library(HDR 1) suffers from color mismatching. To circumvent this problem, we switch one source image in the knowledge library to alter its knowledge distribution. Consequently, the HDR result from the second library(HDR 2) is apparently better than the other, because the misleading patches are reduced and more proper choices are provided. This illustration proves our method's potential to adapt to generalized applications.

\subsection{Swap Methods}
\begin{figure}
\centering
\includegraphics[width=0.49\textwidth]{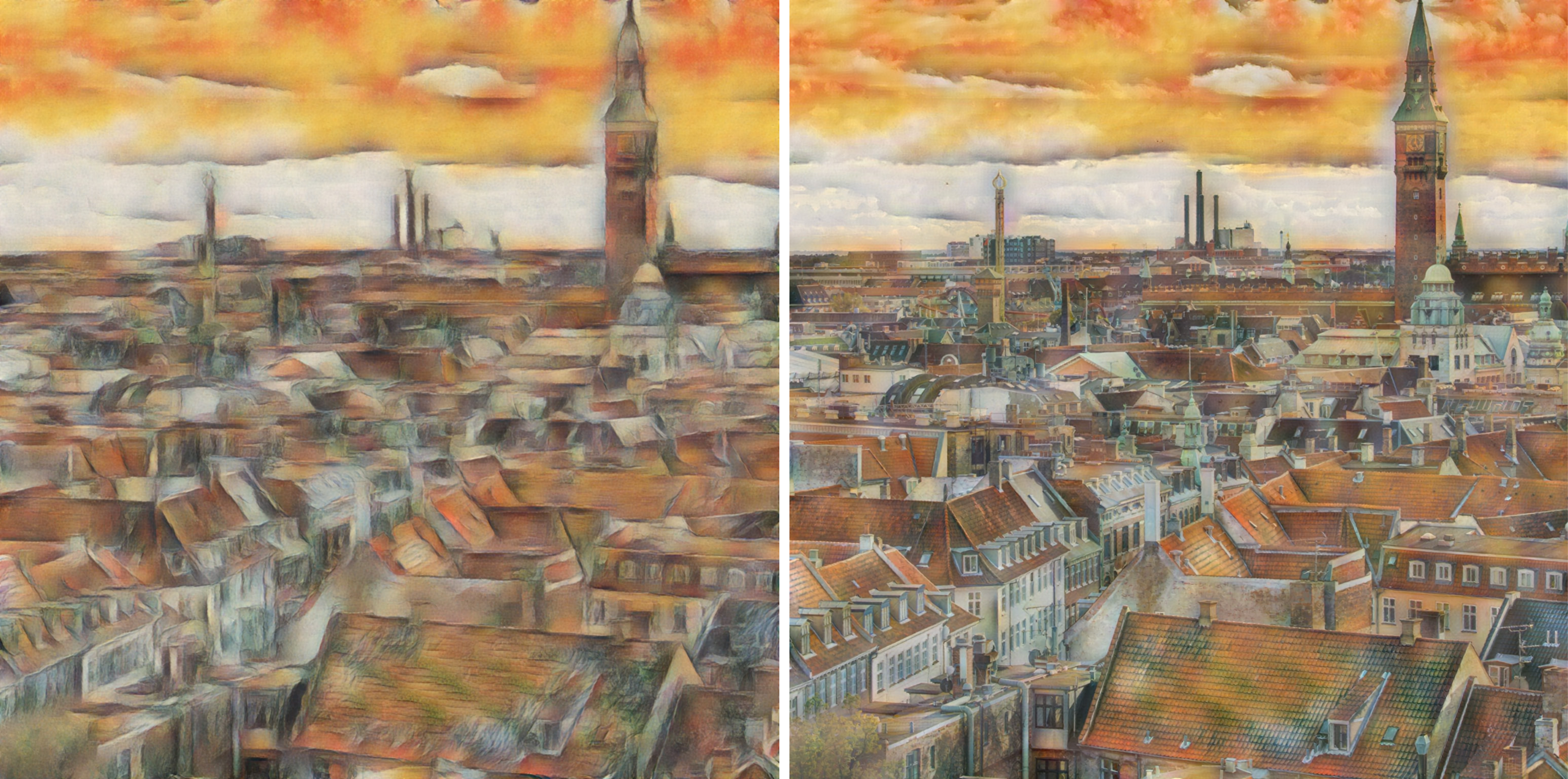}
\renewcommand{\tabcolsep}{42pt}
\begin{tabular}{cc}
Geometric   &  Statistics \\
\end{tabular}
\caption{Stylized images of different swap methods.}
\label{fig:swap}

\centering
\includegraphics[width=0.49\textwidth]{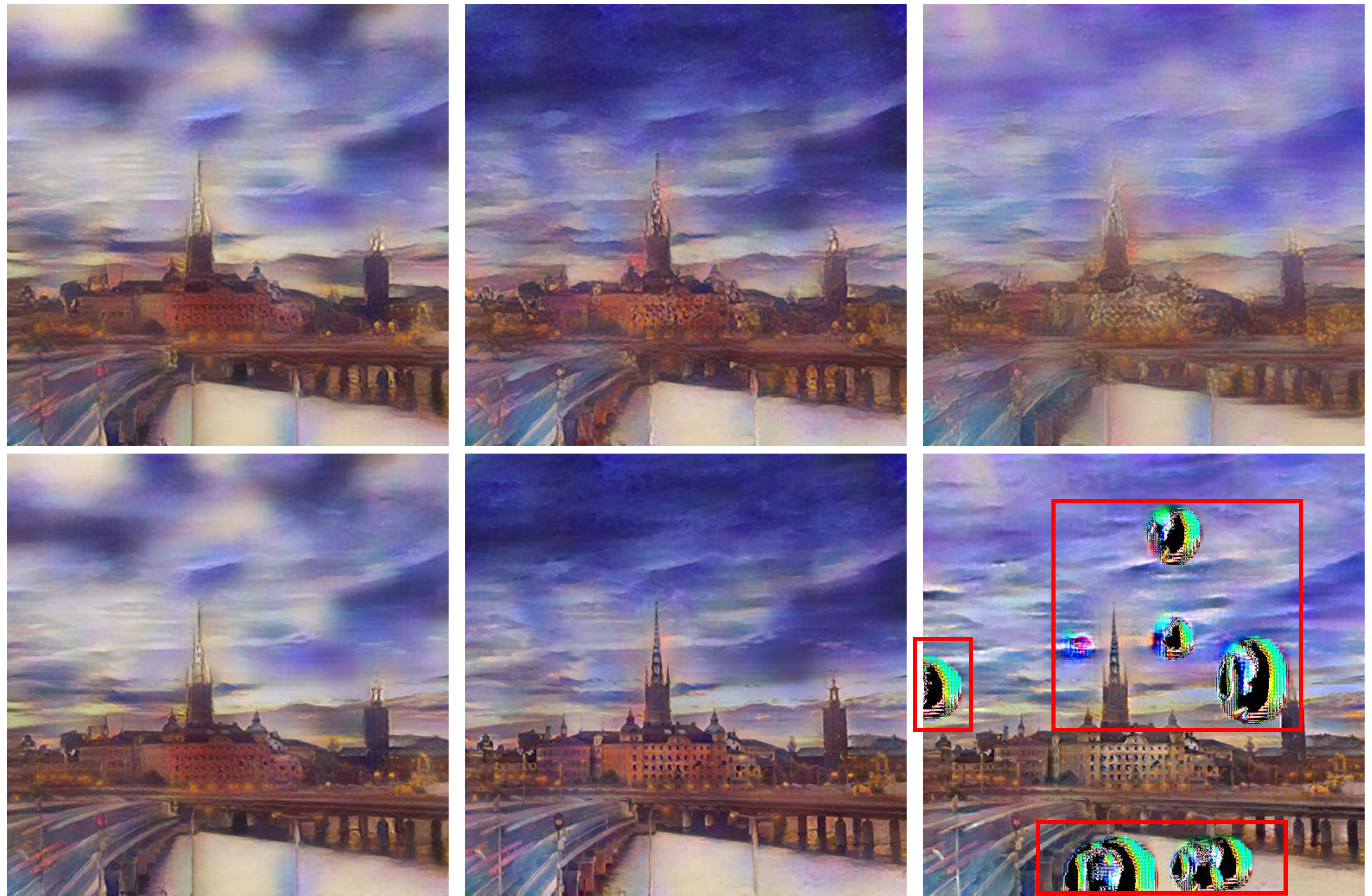}
\renewcommand{\tabcolsep}{29pt}
\begin{tabular}{ccc}
traverse   &  ours  &  PQ \\
\end{tabular}
\caption{Stylized images of both swap methods(above: geometric, below: statistics) by different searching methods. In the marked region are saturation artifacts caused by PQ mismatching.}
\label{fig:search_compare}
\end{figure}

Our framework is able to switch between two means of feature swap: geometric and statistics. They either focus on creating stylistic strokes or preserving semantic details. To compare their effects, we conduct experiments of stylization with same content and styles but different swap methods. The results are shown in Fig.~\ref{fig:swap}. It can be seen that both means have almost equal influence on color range. The contours and shapes, however, are differently. Geometric swap is more suitable for stylization tasks since it excels at altering strokes, while statistics swap can perform better on tasks like HDR by preserving shapes well.

\subsection{Effectiveness of BHKM}
One of the main components of our framework is the BHKM, which is a searching algorithm in nature. In this section, its temporal and spatial complexity is discussed and compared with other searching algorithms that are commonly used for similar purposes. 

\textbf{Temporal Complexity and Efficiency Analysis}
Our method is actually an Approximate Nearest Neighbor (ANN) search method, which tends to achieve a balance between searching accuracy and temporal and spatial complexity.
It avoids massive resource computations of traverse and achieves better searching accuracy than PQ. 
To further conclude the capability of BHKM, we conduct experiments to compare its searching accuracy and time cost with global traverse and PQ. The results are given in Table~\ref{timecost} and Fig.~\ref{fig:search_compare}. We create 9 libraries containing 1-3 images, and run the same content images on each of them to acquire the average time cost.

\begin{table}[]
\renewcommand{\tabcolsep}{12pt}
\begin{tabular}{ccc}
\hline
Approach        & time per patch/s & time per image/s \\ \hline
Traverse        & $3.981\times 10^{-3}$ & $15.799$       \\
PQ              & $3.065\times 10^{-5}$ & $0.122$       \\
ours            & $7.351\times 10^{-5}$ & $0.292$       \\ \hline
\end{tabular}
\caption{Average searching time of different methods on 9 different knowledge libraries with content images of size $512\times512$.}
\label{timecost}
\end{table}

According to these results, our method generally maintains edges and shapes, while more regions in PQ result are blurry or distorted(\eg tower, bridge). In time cost, our method holds great advantage against traverse and is in the same order of magnitude as PQ. In addition, there are obvious saturation artifacts in PQ's statistics swap result. Considering the overall efficiency and effectiveness, ours is the bests among these three searching methods.



\section{Conclusion}

In this paper, we propose an explainable image-translation framework, namely Fine-grained Knowledge Transfer, to achieve image editing with an explicit working mechanism.
In our framework, we are able to accurately understand the formation mechanism of styles, colors, texture patterns, etc.
Contrast to mainstream image translation frameworks,
proposed framework edits the image though geometric pattern matching and imitation.
In order to better match and imitate, we extract and build the target data set into a knowledge library.
The knowledge library is plug $\&$ play for our framework, making the framework can adaptive to different image translation task without training.
Furthermore, to alleviate the search difficulty caused by the data explosion of knowledge library, we have proposed an improved hierarchical K-means search method, BHKM.
We believe this is a very promising image-translation method, because its translation result is just limited by the prior knowledge in the knowledge library, so that the output result is relatively stable.
Experiments also show the feasibility and effectiveness of our framework.
In addition, our method also has its drawbacks: it occupies the local disk to store a large number of patches; the BHKM retrieves an approximate matching target, which causes larger deviation than the traversal method.
In future work, we could concentrate on aforementioned limitations.

{\small
\bibliographystyle{ieee_fullname}
\bibliography{egbib}
}

\clearpage
\appendix

\section*{Appendix: Supplementary Material}
\section{Overview}
In this supplementary material, we first give the detailed analysis of the temporal complexity of our method. Here we conduct more explicit experiments to acquire quantity results. Second, we present the interpretability of our method through a detailed visualization of knowledge backtracking. In the end, we give more high-resolution image translation results to exhibit the performance of our method.
We will release our code on \url{https://github.com/AceSix/Knowledge_Transfer} as soon as possible.

\section{Efficiency Ablation}
Theoretically speaking, we reduce the search space directly by using wavelet to reduce feature dimensions so the overall efficiency is improved. However, it is worth conducting experiments showing the explicit efficiency increase when using different wavelets. So we design experiments to compare the processing speed of Hierarchical K-Means(HKM) with our methods with different degree of dimension reduction. The results are given in Table.~\ref{timecost}. Meanwhile, we randomly choose some generated samples from this experiment to compare the visual effect of each method in Fig.~\ref{fig:time}.

\begin{table}[h]
\renewcommand{\tabcolsep}{12pt}
\begin{tabular}{ccc}
\hline
Approach             & time per patch/s      & time per image/s \\ 
\hline
HKM       & $6.531\times 10^{-5}$ & $0.259$       \\
1-BHKM    & $5.042\times 10^{-5}$ & $0.200$       \\
2-BHKM    & $5.807\times 10^{-5}$ & $0.230$       \\ 
3-BHKM    & $6.564\times 10^{-5}$ & $0.261$       \\ 
4-BHKM    & $7.377\times 10^{-5}$ & $0.293$       \\ 
PQ        & $2.911\times 10^{-5}$ & $0.116$       \\ 
\hline
\end{tabular}
\caption{Average searching time when using a different number of wavelet components for knowledge matching. The numbers in front of BHKM indicate how many wavelet components are used.}
\label{timecost}
\end{table}

\begin{figure*}
\centering
\includegraphics[width=0.98\textwidth]{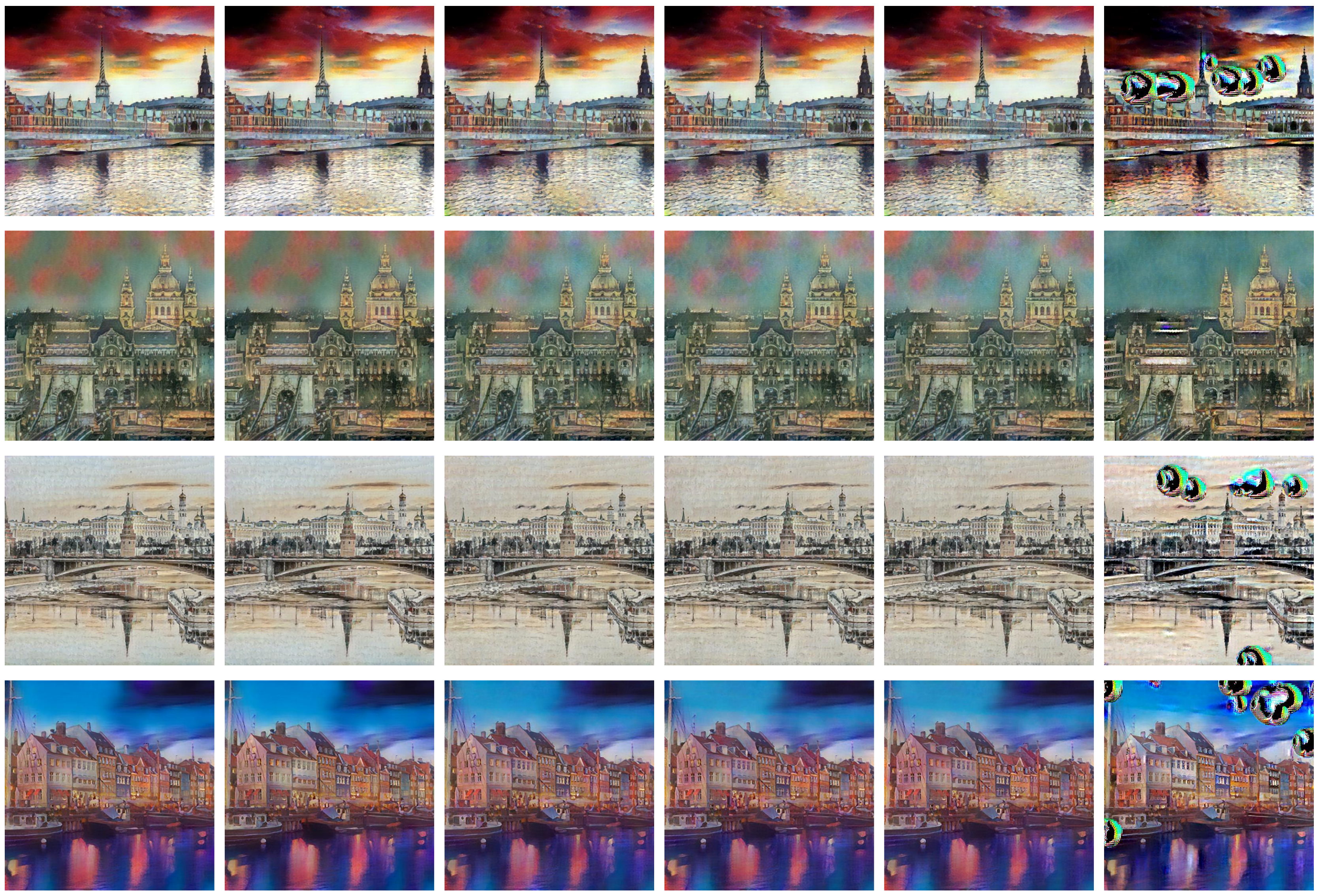}
\setlength{\tabcolsep}{0.047\textwidth}
\begin{tabular}{cccccc}
HKM\quad & 1-BHKM & 2-BHKM & 3-BHKM & 4-BHKM & PQ\quad  \
\end{tabular}
\caption{Stylization results of HKM, BHKM with different numbers of wavelet components and PQ. Please zoom-in the figure for better observation. Please note that the details of the images are partially lost due to file compression.}
\label{fig:time}
\end{figure*}

The results obviously shows the wavelet's role in reducing searching space and increasing efficiency. The time cost brought by wavelet operation is made up when using 3 wavelet components. And using less than 3 wavelet components will give our method an apparent edge against HKM. As the visual comparison shows, the reduction of wavelet components used will not have a perceivable effect on the generation result. Although the time efficiency of PQ surpasses our method, its stylization results mostly contains obvious saturation artifacts.

\section{Knowledge Backtracking}
\begin{figure*}
\centering
\includegraphics[width=0.98\textwidth]{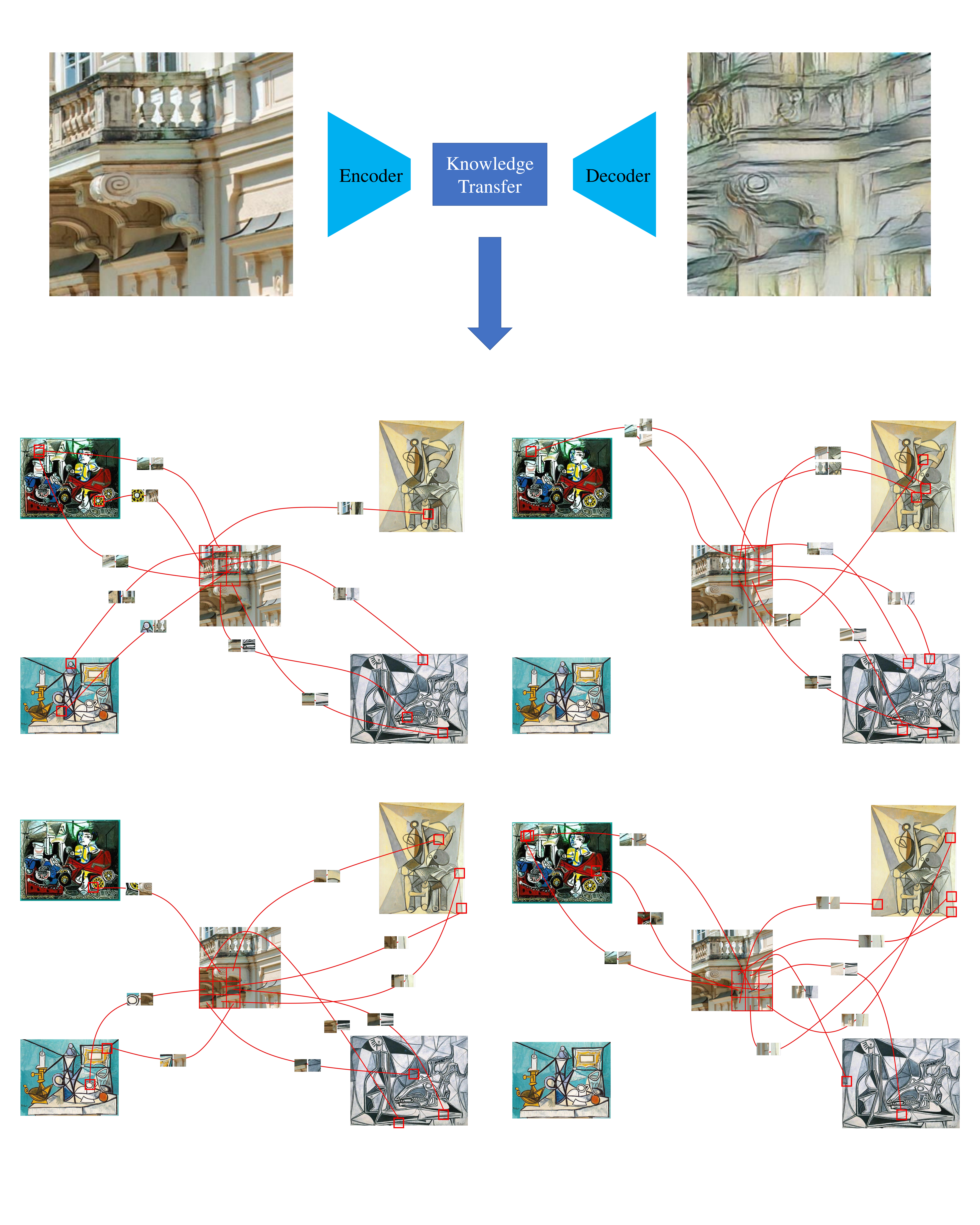}
\caption{Backtracking result of a stylized image. For the sake of clarity, we split all backtracking paths into four graphs. The content patch and its transferred knowledge patch are linked through red lines. The enhanced patch pair is also placed on the corresponding red line. Please zoom-in the figure for better observation. Please note that the details of the images are partially lost due to file compression.}
\label{fig:backtrack}
\end{figure*}

In this section, we explain and demonstrate the interpretability of our method with the help of an explicit graph of backtracking result, Fig.~\ref{fig:backtrack}. This graph gives the knowledge transferred for each patch through a red line from content patch to knowledge patch in its original style image. And we place the enhanced patch pair on each line so their similarity can be directly observed.

\section{Exhibition}
\begin{figure*}
\centering
\includegraphics[width=0.98\textwidth]{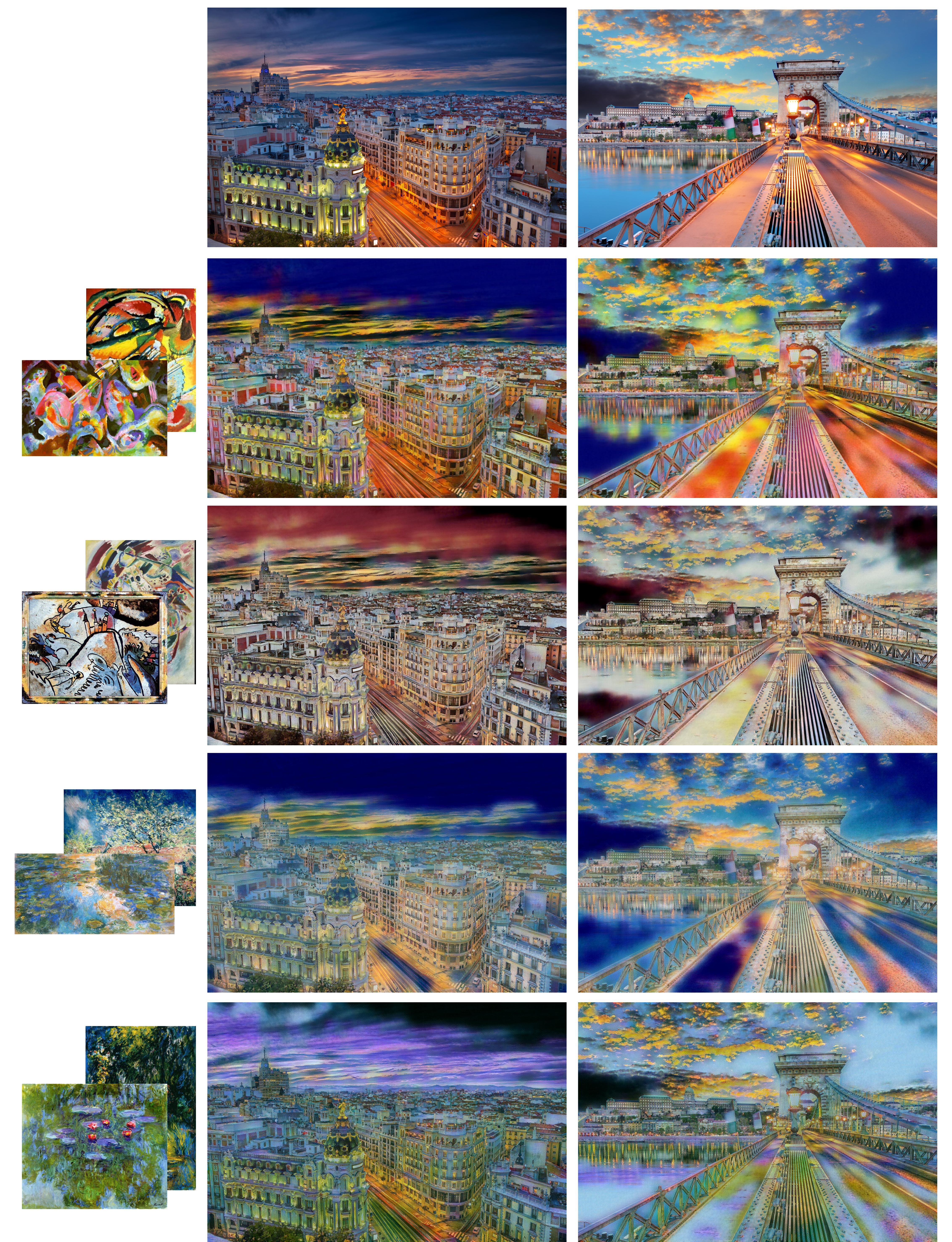}
\caption{Stylization results. The first row and first column are content images and style images in knowledge libraries. Please zoom-in the figure for better observation. Please note that the details of the images are partially lost due to file compression.}
\label{fig:HR2}
\end{figure*}

\begin{figure*}
\centering
\includegraphics[width=0.98\textwidth]{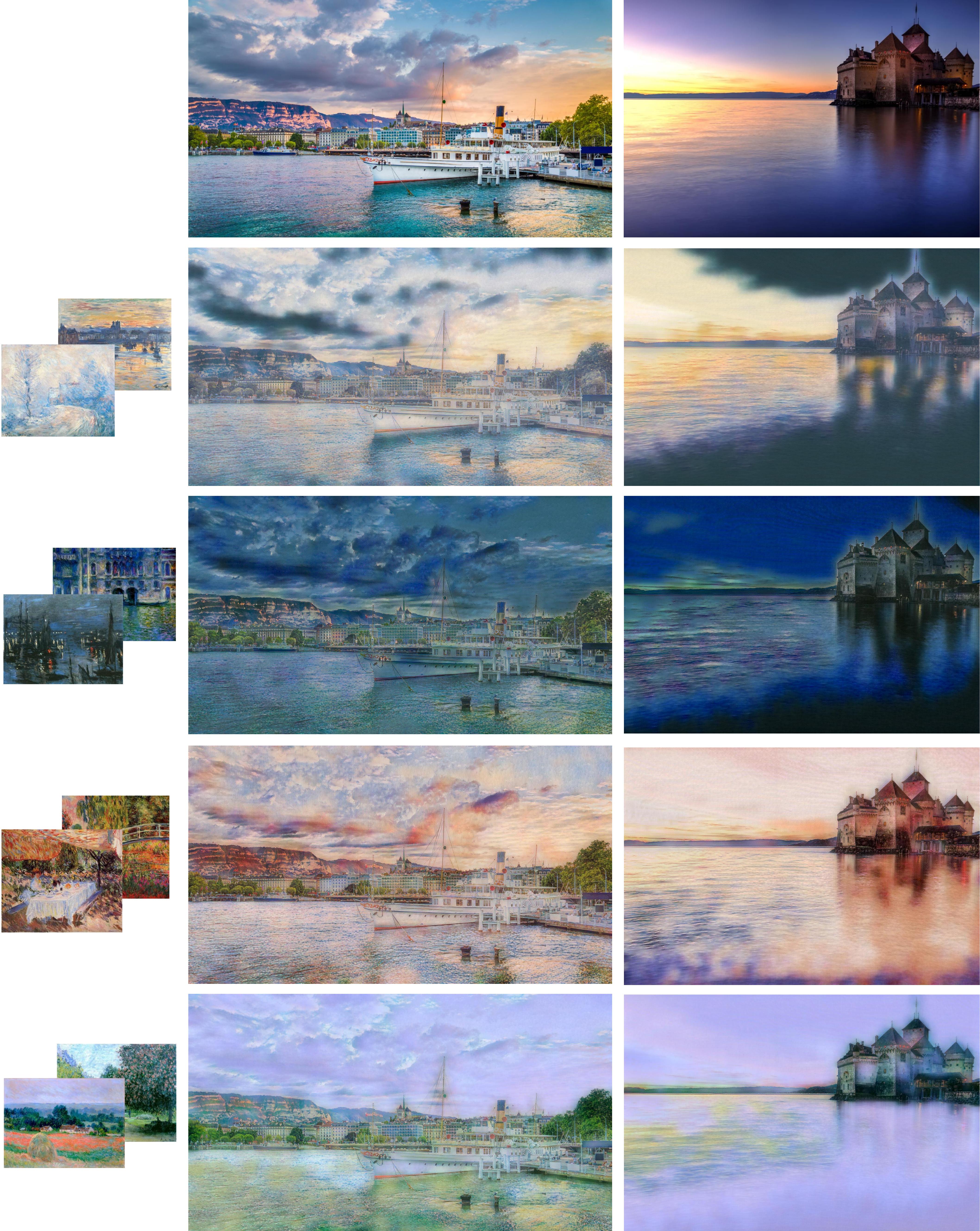}
\caption{Stylization results. The first row and first column are content images and style images in knowledge libraries. Please zoom-in the figure for better observation. Please note that the details of the images are partially lost due to file compression.}
\label{fig:HR3}
\end{figure*}

\begin{figure*}
\centering
\includegraphics[width=0.98\textwidth]{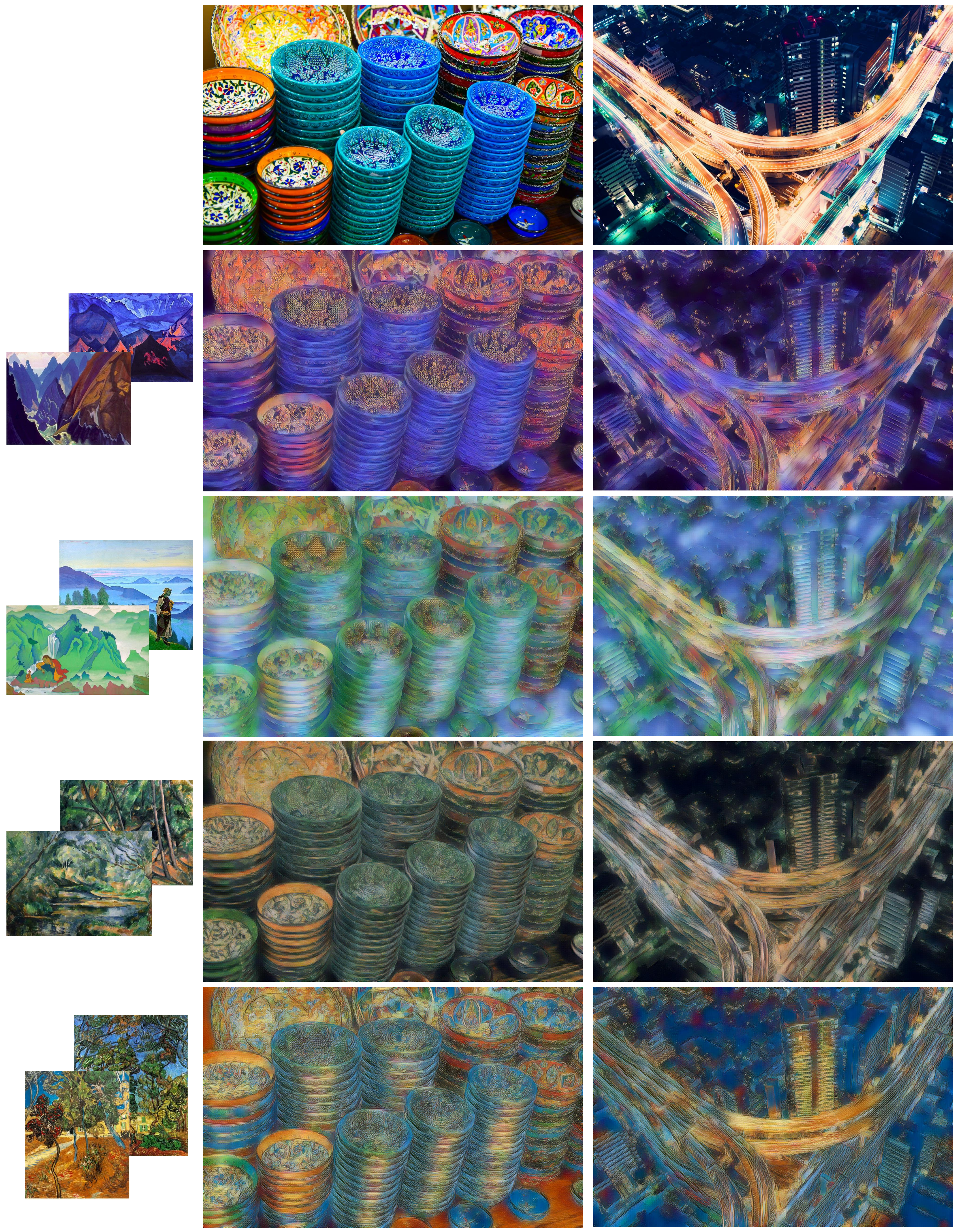}
\caption{Stylization results. The first row and first column are content images and style images in knowledge libraries. Please zoom-in the figure for better observation. Please note that the details of the images are partially lost due to file compression.}
\label{fig:HR4}
\end{figure*}

\begin{figure*}
\centering
\includegraphics[width=0.98\textwidth]{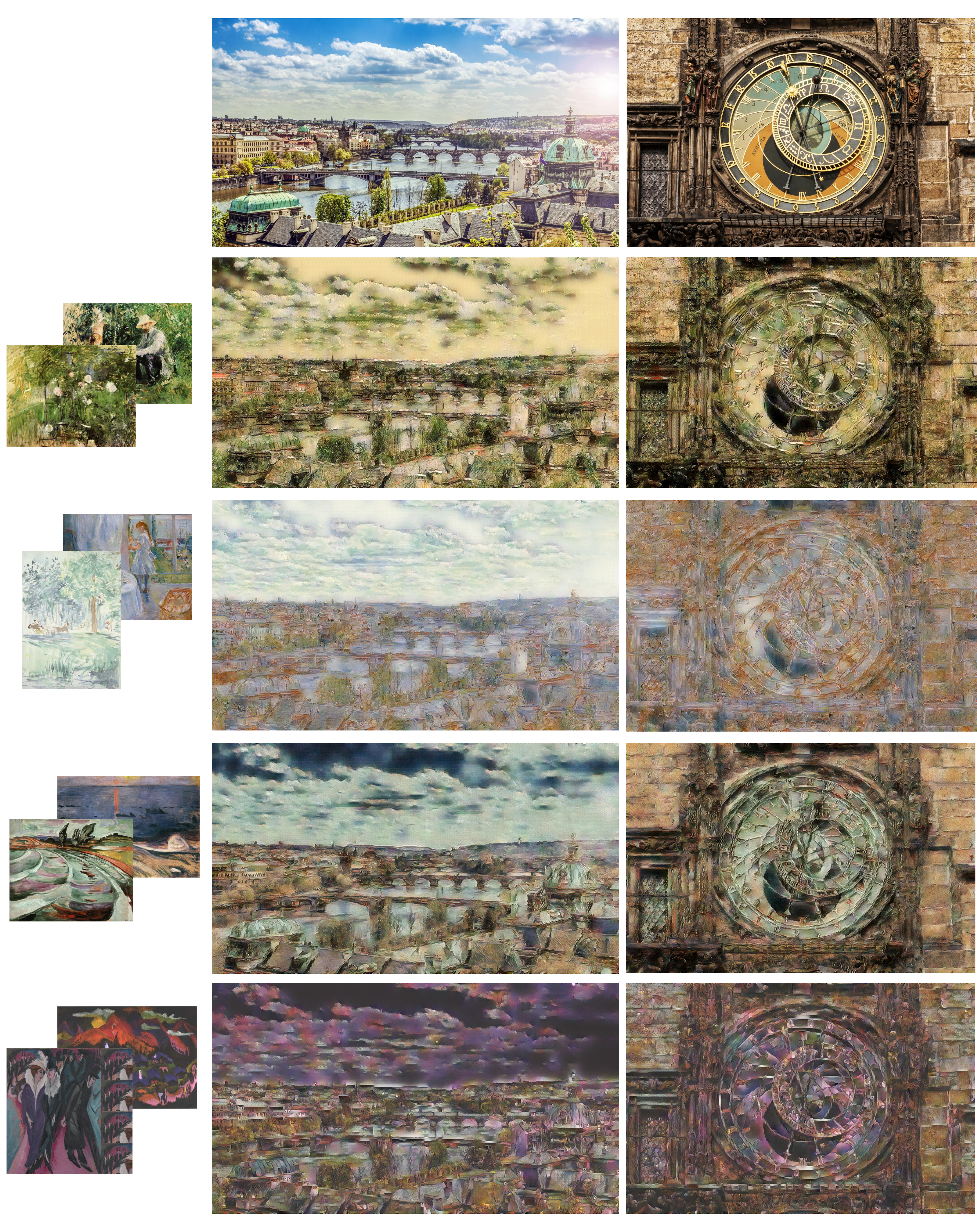}
\caption{Stylization results. The first row and first column are content images and style images in knowledge libraries. Please zoom-in the figure for better observation. Please note that the details of the images are partially lost due to file compression.}
\label{fig:HR1}
\end{figure*}

\begin{figure*}
\centering
\includegraphics[width=0.98\textwidth]{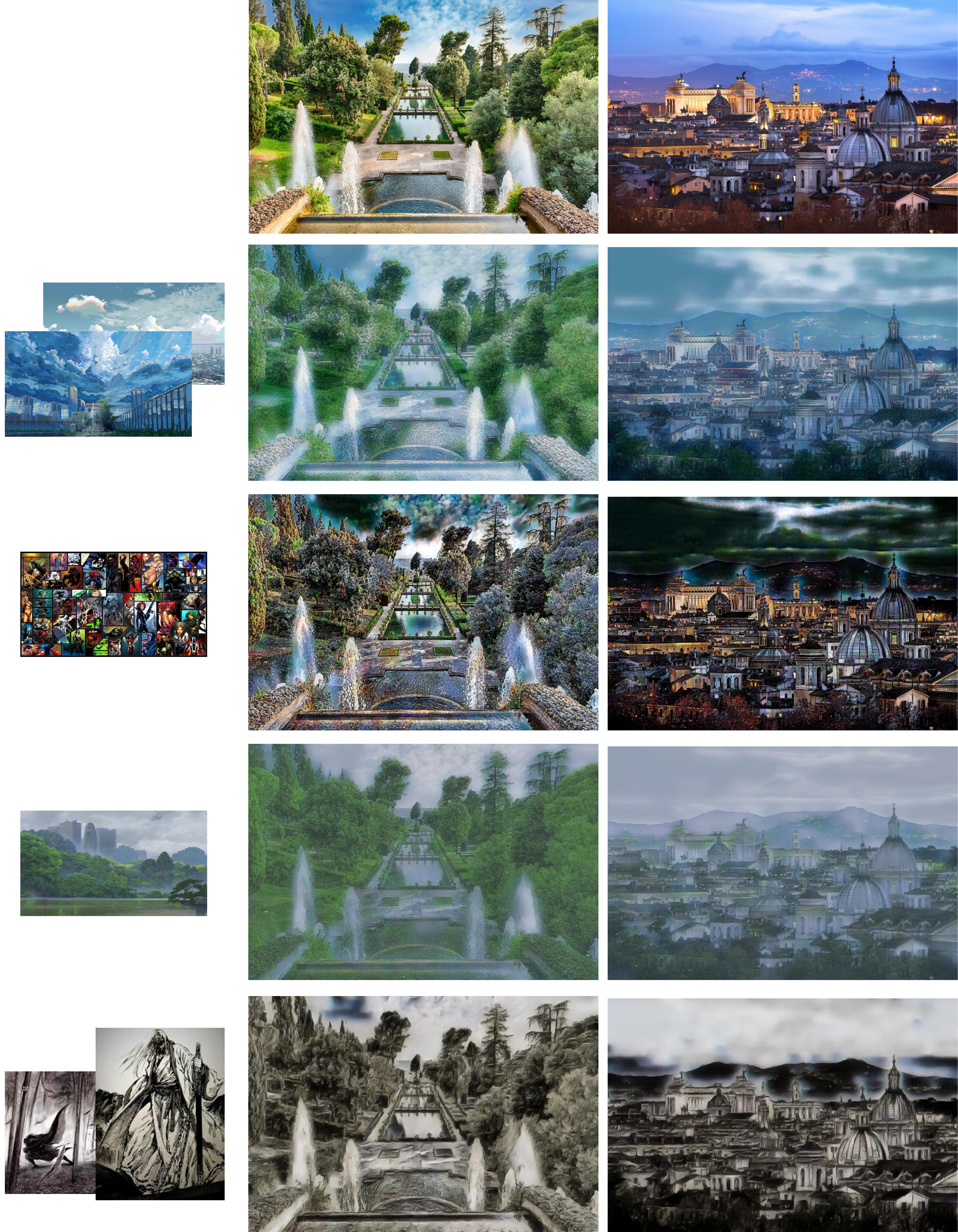}
\caption{Comic Stylization results. The first row and first column are content images and comic images in knowledge libraries. Please zoom-in the figure for better observation. Please note that the details of the images are partially lost due to file compression.}
\label{fig:HR5}
\end{figure*}

\begin{figure*}
\centering
\includegraphics[width=0.98\textwidth]{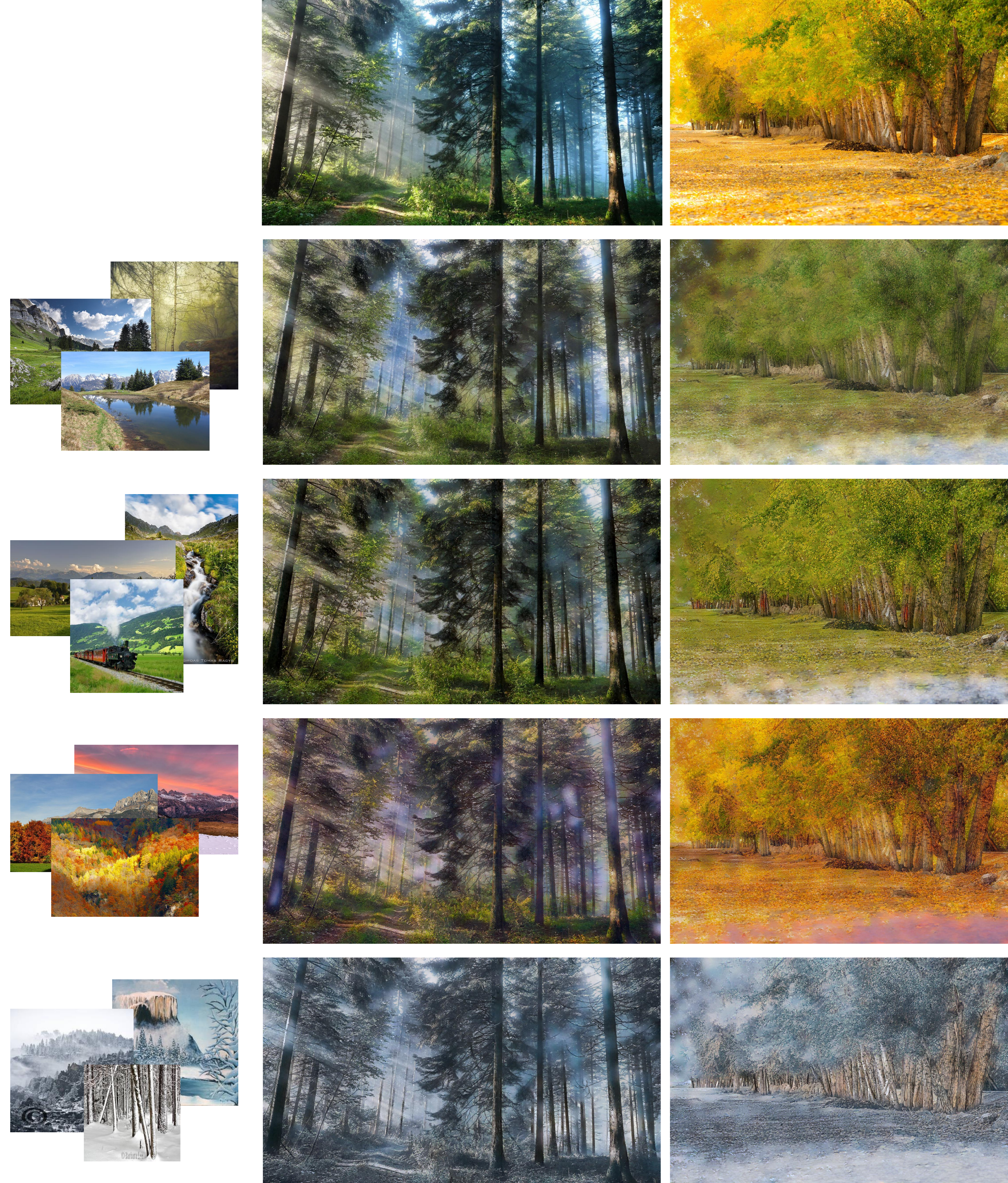}
\caption{Season Transfer results. The first row and first column are content images and season images in knowledge libraries. Please zoom-in the figure for better observation. Please note that the details of the images are partially lost due to file compression.}
\label{fig:HR6}
\end{figure*}

\begin{figure*}
\centering
\includegraphics[width=0.64\textwidth]{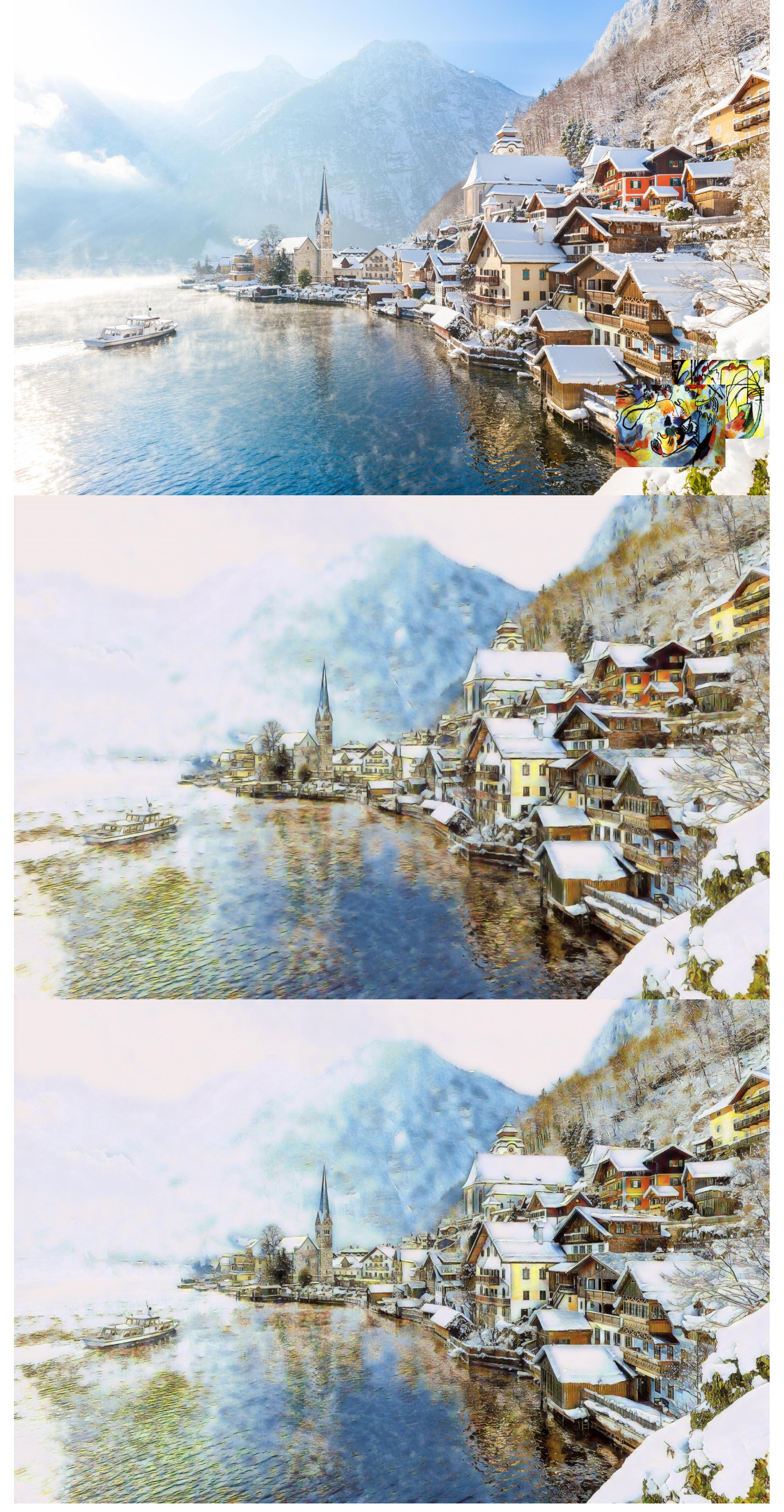}
\caption{Stylization results. Please zoom-in the figure for better observation. Please note that the details of the images are partially lost due to file compression.}
\label{fig:HR7}
\end{figure*}

The plug and play style makes it easy of our method to generate a large quantity of translated images of various tasks. 
The results are given in Fig.~\ref{fig:HR2}, Fig.~\ref{fig:HR3}, Fig.~\ref{fig:HR4} Fig.~\ref{fig:HR1}, Fig.~\ref{fig:HR5}, Fig.~\ref{fig:HR6} and Fig.~\ref{fig:HR7}.

\end{document}